\documentclass[10pt,twocolumn,letterpaper]{article}

\usepackage{cvpr}              

\newcommand{\dname}{\textsc{MMLandmarks}\xspace}

\newcommand{\mypar}[1]{\vspace{0mm}\noindent\textbf{#1}}

\iftrue     
\newcommand{\dimpp}[1]{\textcolor{blue}{[DP: #1]}}
\else
\newcommand{\dimpp}[1]{\textcolor{blue}{\noindent}}
\fi

\usepackage{makecell, multirow, tabularx}
\usepackage{pifont}
\usepackage{subcaption}
\usepackage{amssymb}
\usepackage{amsmath}
\usepackage{graphicx}
\usepackage{booktabs}
\usepackage{tikz,colortbl}
\usepackage{pifont}
\usepackage{dsfont}
\usepackage{svg}
\newcommand{\cmark}{\ding{51}} 
\newcommand{\xmark}{\ding{55}} 
\usepackage[table]{xcolor} 
\definecolor{lightgray}{gray}{0.9}
\usepackage{algorithm}
\usepackage{algorithmic}
\usepackage[accsupp]{axessibility}  

\definecolor{cvprblue}{rgb}{0.21,0.49,0.74}
\usepackage[pagebackref,breaklinks,colorlinks,allcolors=cvprblue]{hyperref}

\def\paperID{***} 
\def\confName{CVPR}
\def\confYear{2026}

\title{\dname: a Cross-View Instance-Level Benchmark \\ for Geo-Spatial Understanding}

\author{
Oskar Kristoffersen$^{1,2}$ \hspace{5mm} Alba Reinders Sánchez$^{1}$ \hspace{5mm} Morten Rieger Hannemose$^{1,2}$ \\ Anders Bjorholm Dahl$^{1,2}$ \hspace{5mm} Dim P. Papadopoulos$^{1,2}$\\
$^{1}$ Technical University of Denmark \hspace{2cm} $^{2}$ Pioneer Center for AI\\
{\tt\small\{ofhkr, albre, mohan, abda, dimp\}@dtu.dk} 
\\ \url{mmlandmarks.compute.dtu.dk}
}

\begin{document}
\maketitle
\begin{abstract}

Geo-spatial analysis of our world benefits from a multimodal approach, as every single geographic location can be described in numerous ways (images from various viewpoints, textual descriptions, geographic coordinates, etc.). 
Current benchmarks have limited coverage across modalities, leading to specialized models that perform well in their respective domains, but do not fully take advantage of other geo-spatial modalities.
We introduce the Multi-Modal Landmark dataset (\textsc{MMLandmarks}), a benchmark composed of four modalities: $197k$ high-resolution aerial images, $329k$ ground-view images, textual information, and geographic coordinates for $18{,}557$ distinct landmarks in the United States. The \textsc{MMLandmarks} dataset has a one-to-one landmark level correspondence across every modality, which enables training and benchmarking models for various geo-spatial tasks, including cross-view Ground-to-Satellite retrieval, ground and satellite geolocalization, Text-to-Image, and Text-to-GPS retrieval.
We show that current specialized and off-the-shelf foundation models cannot be trivially used to solve this variety of geo-spatial tasks, illustrating a gap where multimodal datasets lead to broader geo-spatial understanding. We employ a simple CLIP-inspired baseline that reflects versatility and broad generalization when trained with \dname.
\end{abstract}    
\section{Introduction}
\label{sec:intro}

\begin{figure*}[t]
    \centering
    \includegraphics[width=\textwidth]{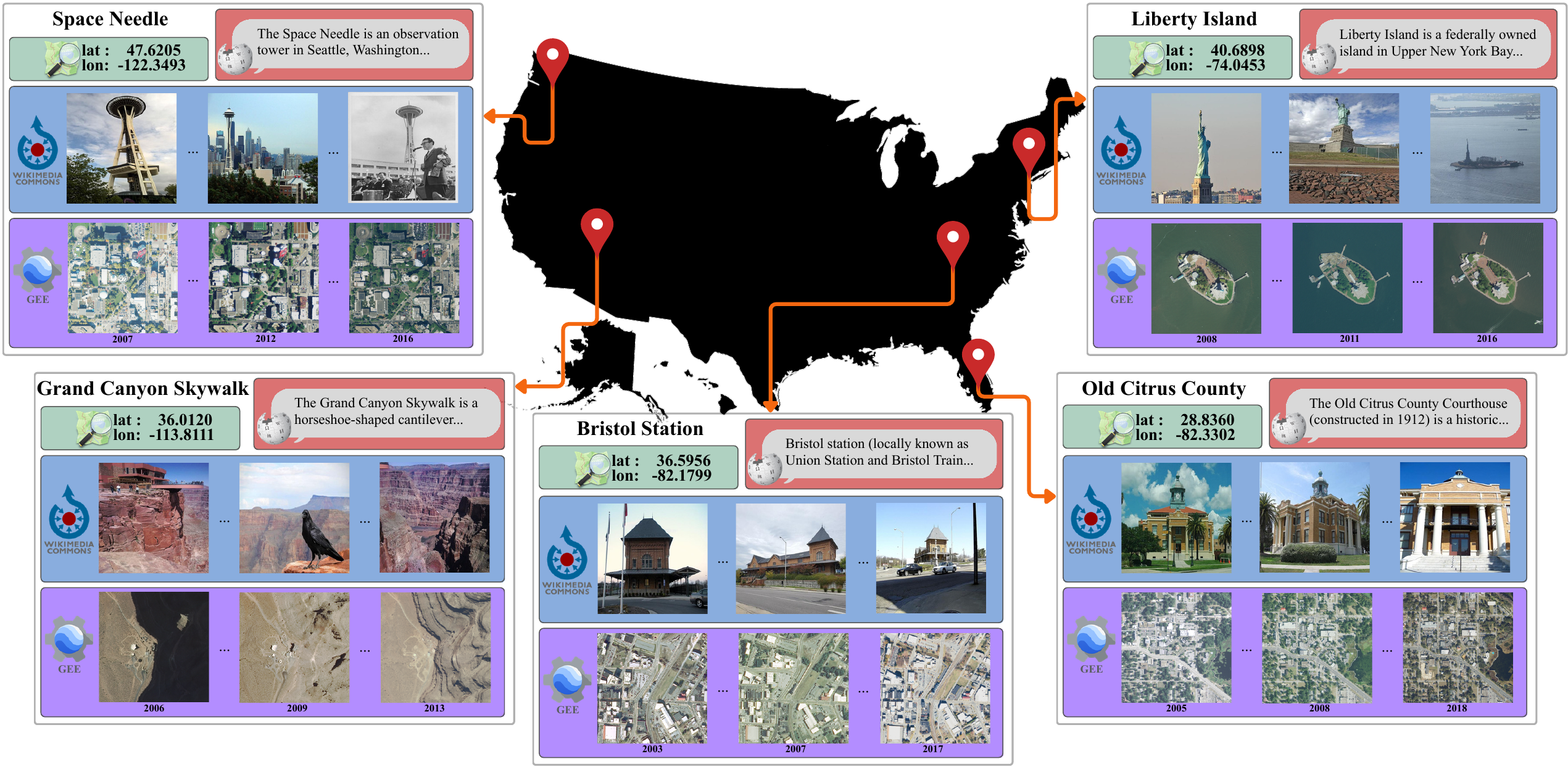}
    \caption{\textbf{\dname.} We present four distinct data modalities: ground-view images, aerial imagery, GPS coordinates, and textual descriptions, collected from $18{,}557$ unique landmarks in the United States. Data sources are included alongside each modality.}
    \label{fig:mml_fig1}
  \end{figure*}

Recent advances in computer vision have been driven by the availability of large-scale datasets for object detection~\cite{lin2014microsoft,deng2009imagenet}, scene understanding~\cite{zhou2017places,xiao2010sun}, and event recognition~\cite{weber2020detecting,weber2022incidents1m}. Similarly, geo-spatial understanding has been motivated by the growing availability of worldwide satellite imagery and a call for remote sensing imagery as a distinct modality in machine learning~\cite{rolf2024mission}. With the vast amount of foundational remote sensing models~\cite{visionfoundationmodelsremote}, many approaches combine remote sensing information from various locations and resolutions. 
Multimodality has become a central part of geo-spatial methods, combining data across visual modalities (e.g.\ multispectral imagery, ground-level views), sensor types (e.g.\ GPS, LiDAR), and additional modalities like text~\cite{wildsat,mall2023remote} or audio~\cite{geobind}. However, many foundational models focus on solving coarse, low-resolution tasks such as landcover semantic segmentation or climate-zone classification, falling short for high-resolution understanding~\cite{akiva2022self,ayush2021geography,cong2022satmae,li2024s2mae,liu2024remoteclip, mall2023remote, mmearth, noman2024rethinking}.
While many aerial and high-resolution satellite datasets exist~\cite{xview, naip, dota, ucmerced}, they are designed for object detection and classification, without providing tasks on an instance level.

Localization is another challenging geo-spatial task where one wants to answer the broad question ``Where in the world was this photo taken?''. Geo-localization models train on large-scale geotagged datasets derived from YFCC100M~\cite{yfcc100m} or Mapillary~\cite{warburg2020mapillary}, intending to directly predict the most accurate GPS location from a query image~\cite{vo2017revisiting, astruc2024osv, clark2023we, haas2023learning,haas2024pigeon,muller2018geolocation, pramanick2022world, weyand2016planet,geoclip}. While effective at regional scales, models struggle with fine-grained localization due to large intra-class variation.
Landmark Retrieval at regional~\cite{oxfordparis} or global~\cite{gldv1, gldv2} levels focuses on exact landmark matching, but is computationally expensive and not scalable. Visual Place Recognition~\cite{berton2023eigenplaces, berton2025megaloc} relaxes the task by classifying images within a certain range of the landmark as positive matches, yet it is still only viable for regional localization.
Cross-view localization bridges the large domain gap between ground and aerial perspectives~\cite{sample4geo, cvcities, cvact, cvusa, vigor, shi2020looking, yang2021cross, zhu2022transgeo, ye2024cross, saig}. However, datasets heavily depend on Google Street View panoramas, which restrict coverage to roads and urban areas. Due to the lack of image diversity, cross-view benchmarks have become saturated. Furthermore, licensing restrictions on satellite and street-view imagery limit the scientific research on datasets collected through Google, slowing research involving street-view and satellite imagery~\cite{helbich2024use}.

Despite the advances of multimodal methods in remote sensing and localization, a core limitation remains: the lack of fine-grained, instance-level datasets that span multiple modalities and viewpoints. Existing datasets fall short in terms of scale, diversity, or annotation granularity, making it challenging to train models capable of reasoning about specific landmarks and locations across modalities. To address these shortcomings, we introduce \dname, a large-scale, instance-level dataset covering $18{,}557$ landmark instances across the United States.

\dname is US-only by design, as the National Agriculture Imagery Program (NAIP)~\cite{naip} is, to our knowledge, the only openly available source of high-resolution aerial imagery with sufficient geographic diversity to support large-scale multimodal geo-spatial research. 
Other open-access remote sensing resources are either low resolution or cover only small, fragmented regions of the world.
~\cref{fig:mml_fig1} shows some examples of landmarks, while ~\cref{fig:collection} illustrates the collection process.
We build a challenging ground-view index set for the retrieval process from the Google Landmarks Dataset v2 (GLDv2)~\cite{gldv2}, from which we remove any overlap with \dname, leading to a gallery of $714k$ ground images. We also collect an aerial index set with $100k$ images, and ensure that there is no overlap with the training set.
\dname is made to push localization as a unified framework where models can be trained and applied to any downstream geo-spatial task. But also to stimulate multimodal learning in a setting where all modalities are available for each instance in the dataset.
To the best of our knowledge, \dname is the first dataset with complete one-to-one landmark-level correspondence across four modalities. It is also the first fine-grained, continental-scale cross-view dataset with permissive licenses that enables training and sharing data and models. We summarize our main contributions as follows:

\begin{itemize}
    \item \textbf{\dname dataset.} We collect a new dataset with complete pairwise correspondence across four modalities: ground-view, aerial-view, text information, and GPS coordinates. The dataset contains $329k$ ground and $197k$ aerial images from $18{,}557$ unique landmarks (see Sec. \ref{sec:dataset}).
    \item \textbf{Geo-spatial multitasking.} Unlike any other specialized geo-spatial datasets, \dname provides a unified framework for training and evaluating models across multiple tasks.
    \item \textbf{Benchmark evaluation.} We report the results of foundational and specialized models on various instance-level geo-spatial tasks and illustrate the potential of training with our dataset by implementing a simple baseline (see Sec. \ref{sec:experiments}).
\end{itemize}

\section{Related Work}
\label{sec:relatedwork}

\begin{table*}[t]
    \caption{\textbf{Dataset comparison.} Modality abbreviations: $S$ - Satellite, $G$ - Ground, $T$ - Text, $C$ - Coordinates, $D$ - Drone. In the Scale column, the number in parentheses indicates the number of cities.}
    \label{tab:dataset-properties}
    \centering
    \resizebox{\textwidth}{!}{
        \begin{tabular}{cccccccccc}
            Task & Dataset & Year & Train(G/S) & Index(G/S) & Instances & Scale (Cities) & Modalities & Open-access & License \\   
            \midrule
            \multirow{5}{*}{\makecell[c]{Geo \\ localization}}
            & IM2GPS \citep{im2gps}           & 2008 & 6.4M/- & - & - & Global & G,C & \xmark & N/A \\
            & YFCC100M \citep{yfcc100m}       & 2016 & 100M/- & - & - & Global & G,C & \cmark & Flickr TC \\
            & PlaNet \citep{weyand2016planet} & 2016 & 126M/- & - & - & Global & G,C & \xmark & N/A \\
            & MP16 \citep{mp16}               & 2017 & 4.7M/- & - & - & Global & G,C & \cmark & Flickr TC \\
            & OSV-5M \citep{astruc2024osv}    & 2017 & 5.1M/- & - & - & Global & G,C & \cmark & CC-BY-SA \\
            \midrule
            \multirow{7}{*}{\makecell[c]{Cross-View \\ Retrieval }} 
            & CVUSA \citep{cvusa}             & 2015 & 35k/35k & 8.8k/8.8k & -  & USA(1) & G,S & \cmark & Flickr TC \\
            & Vo. \citep{vo2016localizing}    & 2016 & 450k/450k & 70k/70k & -  & USA(11) & G,S & \cmark & N/A \\
            & CVACT \citep{cvact}             & 2019 & 44k/44k & 92k/92k & - & Australia(1) & G,S & \cmark & N/A \\
            & Uni-1652 \citep{university1652} & 2020 & 11.6k/701 & 5.5k/1652 & 1652 & 72 Universities & G,S,D & \cmark & N/A \\
            & VIGOR \citep{vigor}             & 2021 & 51k/44k & 53k/46k & -  & USA(4) & G,S & \cmark & N/A \\
            & CV-Cities \citep{cvcities}      & 2024 & 162k/162k & 61k/61k & -  & Global(16) & G,S & \cmark & N/A \\
            & CVGlobal \citep{ye2024cross}    & 2024 & 134k/134k & - & -  & Global(7) & G,S & \cmark & N/A \\
            \midrule
            \multirow{4}{*}{\makecell[c]{Landmark \\ Retrieval}}
            & R-Oxford \citep{oxfordparis} & 2018 & - & 5k + 1M/-  & 11 & Oxford & G & \cmark & Flickr TC/CC \\
            & R-Paris \citep{oxfordparis}  & 2018 & - & 6k + 1M/- & 11 & Paris & G & \cmark & Flickr TC/CC \\
            & GLDv1 \citep{gldv1}          & 2018 & 1.2M/- & 1.1M/- & 30k & Global & G & \xmark & Multiple \\
            & GLDv2 \citep{gldv2}          & 2020 & 4.1M/- & 762k/- & 200k & Global & G & \cmark & CC/Public-domain \\
            \midrule
            \rowcolor{lightgray}
            & \textbf{\dname} & 2026 &  329k/197k & 714k/100k & 18,557 & USA & G,S,T,C & \cmark & CC/Public-domain \\
            \bottomrule
        \end{tabular}
    }
\end{table*}

\mypar{Multimodal Learning.}
CLIP~\cite{radford2021} sparked recent progress in multimodal learning, followed by supervised~\cite{girdhar2022omnivore, likhosherstov2021polyvit} and self-supervised~\cite{arandjelovic2017look, imagebind, girdhar2023omnimae, morgado2021audio, tian2020contrastive} approaches for learning joint representations with two or more distinct modalities~\cite{alayrac2020self,baevski2022data2vec,imagebind,guzhov2022audioclip,nagrani2022learning,languagebind}. ImageBind~\cite{imagebind} and LanguageBind~\cite{languagebind} align multiple modalities with image and text anchors, enabling cross-modal associations with no direct supervision.
Multimodal representation learning requires large-scale paired datasets, often web-scraped from the internet~\cite{yfcc100m,imagebind,radford2021}. Most existing models use image-text pairs, while one-to-one data alignment across multiple modalities remains limited.

In geo-spatial understanding, multimodal learning is also applied for various tasks~\cite{geoclip,wildsat,geobind,satclip}. WildSat~\cite{wildsat} and GeoBind~\cite{geobind} use ecological markers to obtain representations of animal habitats~\cite{birdsat,taxabind,stevens2024bioclip}, while
SatCLIP~\cite{satclip} and GeoCLIP~\cite{geoclip} leverage geo-tagged ground and aerial imagery for localization.
Other efforts rely on open-source multi-spectral satellite and aerial imagery~\cite{akiva2022self, ayush2021geography, cong2022satmae, li2024s2mae, li2024masked, liu2024remoteclip, mall2023change, manas2021seasonal, mmearth, noman2024rethinking, wang2022rsp},
or incorporate language supervision to enhance semantic understanding in remote sensing~\cite{luo2024skysensegpt, mall2023remote, shabbir2025geopixel, xu2024rs}.
In the geo-spatial domain, proposed methods~\cite{satlas,francis2024major,skysense,zhang2024rs5m} combine remote sensing datasets with aerial~\cite{resisc45, christie2018functional, xview, maxwell2017naip, sun2022fair1m, aid, dota, ucmerced, isaid, patternnet}, and satellite images~\cite{long2021creating, helber2019eurosat, schmitt2019sen12ms, bigearthnet, sumbul2021bigearthnet}.
%
Unlike most existing models limited to specific modality pairs, \dname supports dense, instance-level supervision across all modality combinations, enabling the learning and evaluation of truly unified geo-spatial models.

\mypar{Image-to-Image Localization.}
Localization aims to determine where a query image was taken through retrieval. Early approaches rely on image retrieval techniques with Ground-to-Ground retrieval~\cite{oxfordparis,im2gps,arandjelovic2016netvlad,gldv1, gldv2, yokoo2020two, HGLDv2, berton2025megaloc}.
More recent work addresses the localization task by bridging the domain gap between ground and aerial images through cross-view retrieval~\cite{shi2019spatial, shi2020looking, yang2021cross, zhu2022transgeo, saig, sample4geo, zhang2024geodtr+}, with multiple benchmarks sampling image pairs from various cities and regions~\cite{cvusa, vo2016localizing, cvact, ye2024cross, vigor, cvcities, university1652}.
To address the perspective change between ground-level panoramas and overhead satellite imagery, many state-of-the-art methods employ geometric transformations or warping techniques, aligning one view with the other~\cite{shi2019spatial,ye2024cross,li2024unleashing}. 
Cross-view benchmarks have become saturated due to the strong geometric correspondence between views and the lack of diversity in image content, which fails to reflect the complexity and variability of real-world scenes~\cite{sample4geo}.
In contrast, \dname addresses these shortcomings by adopting a landmark-centric approach. Rather than relying on road-based panoramas or coarse regional alignment, \dname provides more variation across ground and aerial views by collecting landmark information from OpenStreetMaps~\cite{openstreetmap} and Wikimedia Commons.

\mypar{Geolocalization.}
This task reformulation addresses the infeasibility of Image-to-Image localization at a global scale, by discretizing the Earth into geo-cells~\cite{weyand2016planet, vo2017revisiting, seo2018cplanet, muller2018geolocation, pramanick2022world, haas2024pigeon}.
To incorporate information from additional modalities, \citet{haas2023learning} uses CLIP's text encoder to iteratively predict the most likely country and city, while \citet{geoclip} combines CLIP's image encoder with a GPS encoder. 
Recent works utilize LLMs and VLMs to further improve geolocalization~\cite{li2024georeasoner, jia2025georanker, jia2024g3}, or to explore downstream tasks~\cite{campos2026gaea}.
%
%
\dname enables the development and evaluation of models to overcome the ambiguity of large intra-class variation by incorporating overhead views of all locations in the training process.

\section{MMLandmarks Dataset}
\label{sec:dataset}

\begin{figure*}[t]
    \centering
    \includegraphics[width=\textwidth]{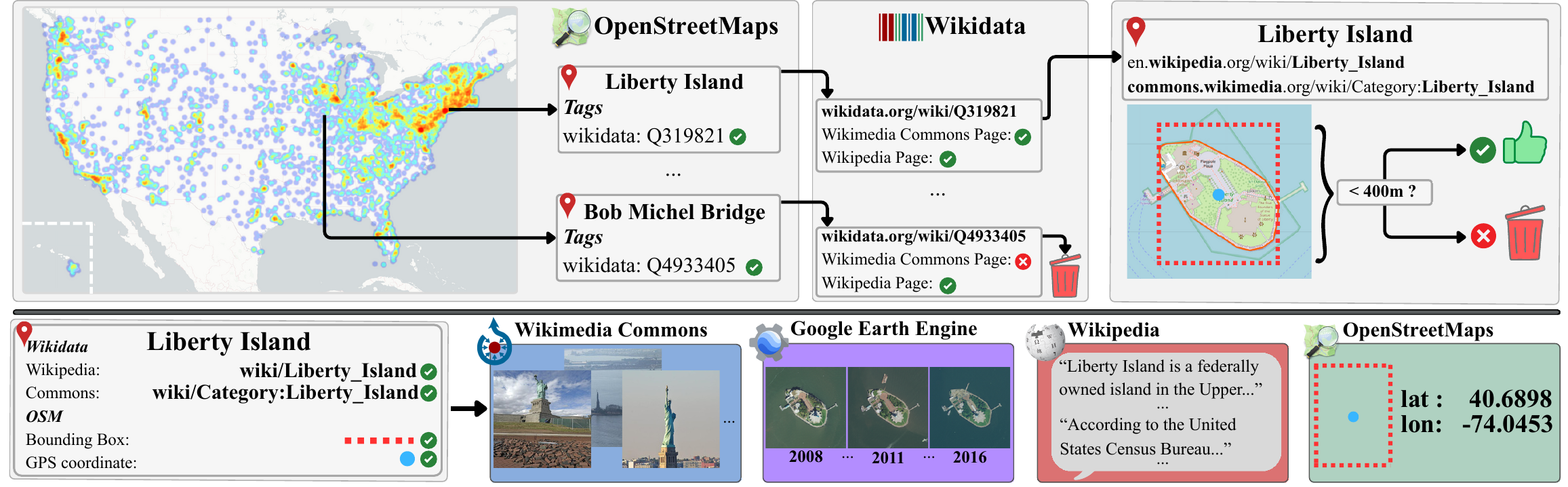}
    \caption{\textbf{Pipeline for collecting the landmarks with the required criteria.} Tags from OpenStreetMaps are used to collect Wiki-identifiers, ensuring that landmarks have a Wikipedia and Wikimedia Commons page. If both are available, we check that the longest edge of the landmark's bounding box is smaller than 400 meters to keep an even size distribution across the dataset. Every resulting landmark has a Wikimedia Commons page (ground), a Wikipedia page (text), a box size and center (coordinates), and associated aerial imagery (satellite).}
    \label{fig:collection}
\end{figure*}

\dname allows systematic investigation of methods for accurate location prediction using multiple information sources, a capability that current datasets only partially support (see Tab.~\ref{tab:dataset-properties}). With more high-resolution satellite images, cross-view localization becomes a critical step towards unified and globally scalable visual understanding. Combining coordinate prediction from geolocalization models with fine-grained cross-view retrieval in designated areas offers a way to solve two problems: accurate localization and retrieval at a global scale.

Each data point (landmark) in \dname has several ground view images and temporal aerial images, a Wikipedia text corpus, and GPS coordinates. The dataset exhibits several realistic properties: 
\textit{Intra-class variability}: The photos collected from Wikimedia Commons inherently contain high variations in lighting conditions, camera angles, and viewpoints, including indoor and outdoor images of the landmarks. The images may also depict other elements indirectly related to the landmarks, such as drawings or scanned documents.
\textit{Continent scale}: The dataset offers instance-level cross-view images from the United States.
\textit{Long-tailed class distribution}: The number of images per landmark and their geographic location vary significantly depending on the popularity of the landmark.
\textit{Temporal change}: The aerial imagery of the NAIP enables the collection of high-resolution images of landmarks at various timestamps, providing an additional complexity and a natural augmentation to the dataset, which is rarely found in current remote-sensing datasets.


\subsection{Landmark Collection}
Finding associations between ground and satellite views of the same instances is a complex task. A straightforward approach is to use Google Maps and Street View to sample satellite views and road panoramas~\cite{cvusa, vo2016localizing, cvact, vigor, cvcities, ye2024cross}. This cannot be trivially adapted to natural geo-tagged images, since images of a landmark are typically taken from a distance. When sampling the corresponding satellite view, we may end up with no landmark at all since the ground image was taken too far away. Inspired by GLDv2~\cite{gldv2}, we leverage the landmarks available in Wikimedia Commons to create a trustworthy correspondence between the ground views and aerial views in \dname. 

We start by collecting the meta-information of all polygons in the United States from OpenStreetMaps~\cite{openstreetmap}, and keep all those containing at least one of the tags \{\texttt{wikipedia} $\cup$ \texttt{wikidata}\}. Thereafter, we collect the Wiki-identifier for each polygon from its associated Wikipedia and Wikidata pages. 
The Wiki-identifier has the format "Q123456", and links all Wiki-information to a single entity. We filter the landmarks and check for the presence of a Wikipedia page containing textual information, and a Wikimedia Commons page containing ground view images. 
As a final step, to ensure an even size distribution across all satellite images, we heuristically decide to only keep landmarks where the largest side of their bounding box is less than 400 meters. 
After the filtration process (see \cref{fig:collection}), we obtain $18{,}557$ landmarks in the US with all four modalities. Each landmark has a varying number of ground and aerial-view images, and a single GPS coordinate and Wikipedia text. \dname therefore contains $329{,}349$ ground images, $197{,}205$ aerial images, $18{,}557$ GPS coordinates and $18{,}557$ textual corpora.

\subsection{Data Collection}

In retrieval-based tasks, the goal is to rank images from a large index set according to their similarity with a given query. We construct both a dedicated training set, a diverse index set, and a query set.

\mypar{Training set.} The ground images are collected from Wikimedia Commons, which is a large Media Repository containing millions of images licensed under Creative Commons and Public Domain licenses. The images are primarily provided by photographers and organizations, most of which come from yearly Wiki Loves Monuments events. Similar to GLDv2~\cite{gldv2}, we resize the images such that the largest side is a maximum of $800$ pixels.

We sample aerial images from the publicly available NAIP, which contains aerial images of the United States with a pixel size of one to two meters. We use Google Earth Engine to collect the images, which provides data for non-commercial and research purposes. For each location, we collect aerial images of size $800\times800$ from the years in which they are available. The aerial imagery contains variations (natural augmentations) resulting from differences in time, weather, angle, sun orientation, and vegetation.

The GPS coordinates of all landmarks are set as the center of the landmark, and all text sections from the landmarks' Wikipedia pages are collected, while discarding sections such as "References" or "See also".

\mypar{Index set.} We create two large index sets to evaluate models on the different geo-spatial tasks. 
We use the ground index set from GLDv2~\cite{gldv2}, which contains $762k$ images of $101k$ landmarks across the world. We filter the landmarks in the index set by country, and find $17{,}804$ belonging in the US. Since the images are sampled similarly to GLDv2~\cite{gldv2}, we remove the $5{,}277$ landmarks overlapping with the \dname dataset from the index set, yielding a final ground index set of $714{,}554$ images.
To avoid a large domain gap between the aerial training and index set, which would make Ground-to-Satellite retrieval easier, we randomly sample landmarks from the training set and add some noise to their GPS coordinates until obtaining $100k$ new locations. After ensuring that the new coordinates are more than 500 meters away from any other coordinate in the training set (to avoid seeing a landmark in other images) and any of the other index locations, we sample the latest aerial NAIP image from these coordinates.

\mypar{Query set.} We randomly sample $1000$ landmarks from the training set. For all landmarks, ground images exhibit high intra-class variability, making the benchmark challenging. However, the aerial images from each landmark are highly correlated, which could inflate performance. We therefore use all ground-views from each landmark as queries ($18{,}688$ images), but only the latest aerial-views ($1000$ images). We keep the $1000$ GPS coordinates and textual corpora untouched. The resulting tasks all have a single ground truth to be correctly retrieved, except for \textit{X}-to-Ground retrieval, where multiple ground truths can be found.

\subsection{Data processing}

Since the ground images are crowd-sourced, they contain both indoor and outdoor views of the landmark. However, indoor images may not be appropriate for learning geo-spatial aligned representations.
We employ a Vision Language Model (\texttt{llava-hf/llava-1.5-7b-hf})~\cite{llava} and prompt it to categorize every image as indoor or outdoor. We make a subset of the \dname training set where only the outdoor images are used, which corresponds to 83\% of the original ground views ($259{,}451$ outdoor against $51{,}210$ indoor images). To verify the categorization accuracy, $1000$ randomly sampled images are reviewed. Only $8.2\%$ of the images are wrongly labeled, most of which are zoomed in or without meaningful context.
Further examples and details about the filtering process are provided in the Supplementary Material.
In \cref{sec:experiments}, both the full dataset and the subset are considered during training. Unless stated otherwise, the full dataset is used.
\begin{table*}[t] 
    \caption{\textbf{Cross-view ground and satellite retrieval.} Comparison of median rank (medR, lower is better), mean Average Precision at $1000$ (mAP@1k, higher is better), and recall at $K$ (R@$K$, higher is better) between off-the-shelf cross-view Ground-to-Satellite retrieval models and multimodal foundational models on \dname. Median rank is the median position where the first correct match is retrieved. The query and index sizes for satellite to ground are $1000$ and $733k$, respectively, and for ground to satellite are $18{,}688$ and $101k$.}
    \label{tab:ground_sat_table}
    \centering
    \resizebox{\textwidth}{!}{
        \begin{tabular}{cccc|ccccc|ccccc}
            \toprule
             & \multirow{2}{*}{\makecell{model}} & \multirow{2}{*}{\makecell{arch}} & \multirow{2}{*}{\makecell{train res}} & \multicolumn{5}{c}{\makecell{Satellite $\rightarrow$ Ground}}  & \multicolumn{5}{c}{\makecell{Ground $\rightarrow$ Satellite}}  \\
            & & & & medR $\downarrow$ &  mAP@1k$\uparrow$  & R@1 $\uparrow$ & R@5 $\uparrow$ & R@10 $\uparrow$ & medR $\downarrow$ & mAP@1k$\uparrow$ & R@1 $\uparrow$ & R@5 $\uparrow$ & R@10 $\uparrow$ \\
            \midrule
            \multirow{15}{*}{\rotatebox{90}{\textbf{\textit{Zero-Shot}}}} 
                & DINO~\cite{dino}               & ViT-B & 224 & 84050 & 0.3 & 1.3 & 2.5 & 3.6 & 27836 & 0.6 & 0.3 & 1.4 & 2.2 \\
                & SigLIP~\cite{siglip}           & ViT-B & 256 & 4801 & 0.8 & 2.9 & 7.4 & 10.3 & 5437 & 5.9 & 4.0 & 12.0 & 17.3 \\ 
                & SigLIP~\cite{siglip}           & ViT-B & 384 & 3854 & 1.2 & 2.7 & 8.7 & 12.8 & 4051 & 7.6 & 5.3 & 15.2 & 20.9 \\   
                & SigLIP~\cite{siglip}           & ViT-B & 512 & 5110 & 1.1 & 2.8 & 8.6 & 12.7 & 4387 & 8.0 & 5.8 & 15.2 & 20.6 \\ 
                & OAI-CLIP~\cite{radford2021}    & ViT-B & 224 & 1695 & 1.4 & 4.0 & 11.3 & 15.8 & 2215 & 7.9 & 5.8 & 15.2 & 22.0 \\
                & SigLIP2~\cite{siglip2}         & ViT-B & 512 & 2461 & 1.8 & 4.8 & 10.8 & 14.4 & 431 & 11.7 & 8.5 & 21.5 & 28.6 \\ 
                \cmidrule(lr){2-14}
                & DiNOv2~\cite{dinov2}           & ViT-L & 518 & 28342 & 0.7 & 1.9 & 3.0 & 4.2 & 9335 & 4.7 & 3.3 & 8.8 & 12.5 \\ 
                & DiNOv3~\cite{dinov3}           & ViT-L & 518 & 14623 & 1.5 & 3.7 & 7.4 & 10.3 & 2241 & 8.5 & 6.1 & 15.5 & 22.3 \\
                & SigLIP~\cite{siglip}           & ViT-L & 256 & 2602 & 2.0 & 5.3 & 11.7 & 16.7 & 1302 & 10.8 & 8.0 & 19.1 & 25.6 \\ 
                & SigLIP~\cite{siglip}           & ViT-L & 384 & 1546 & 3.0 & 7.2 & 15.2 & 19.5 & 679 & 13.8 & 10.7 & 23.8 & 30.8 \\   
                & OAI-CLIP~\cite{radford2021}    & ViT-L & 336 & \underline{519} & \underline{4.9} & \underline{10.4} & \underline{21.2} & \underline{27.5} & 620 & 15.2 & 11.9 & 26.4 & 35.2 \\  
                & SigLIP2~\cite{siglip2}         & ViT-L & 512 & 682 & 4.1 & 8.6 & 18.6 & 23.4 & \underline{140} & \underline{18.7} & \underline{14.6} & \underline{31.6} & \underline{40.1} \\ 
                \cmidrule(lr){2-14}
                & Uni-1652~\cite{university1652}       & R50 & 256 & 62606 & 0.1 & 0.4 & 0.8 & 1.3 & 39961 & 0.4 & 0.2 & 0.9 & 1.7 \\
                & TransGeo-$90^{\circ}$ FoV~\cite{zhu2022transgeo} & DeiT & 256 & 40973 & 0.1 & 0.7 & 2.0 & 2.9 & 13425 & 0.9 & 0.4 & 1.7 & 3.1 \\
                & Sample4Geo-UNI~\cite{sample4geo} & ConvNext-B & 384 & 34988 & 0.4 & 3.0 & 6.2 & 8.0 & 40056 & 1.1 & 0.85 & 2.8 & 4.5\\
            \midrule
            \rowcolor{lightgray}
            & \textbf{MMCLIP}   & CLIP (ViT-L) & 336 & \textbf{23} & \textbf{18.8} & \textbf{30.4} & \textbf{52.4} & \textbf{61.3} & \textbf{48} & \textbf{26.2} & \textbf{20.5} & \textbf{46.5} & \textbf{58.4} \\ 
            \bottomrule
        \end{tabular}
}
\end{table*}

\subsection{Distribution and Challenges}

The \dname dataset inherits multiple relevant and realistic challenges from the image collection through Wikimedia Commons and the NAIP. 
\textit{Skewed distributions}: The landmarks collected are biased in terms of geographical locations and category. As commonly seen in other localization datasets~\cite{yfcc100m,mp16}, most images posted online are taken in well-known locations, which is also reflected in our dataset (see \cref{fig:collection}). There are large concentrations of landmarks around major cities, especially in California and the Northeast. 
%
\textit{Temporal change}: We take advantage of the NAIP to sample aerial images at multiple timestamps, sometimes spanning over a decade. As a consequence, changes may occur both for the landmark and its surroundings. This opens for research in temporal change detection.
\textit{Large domain gap}: Learning the correspondence between ground and aerial imagery can become difficult due to large perspective changes from one modality to another. The large intra-class variability of ground images makes the task harder, as landmark images are taken from various viewpoints, including both indoor and outdoor settings.

\subsection{Licenses}
All images in \dname are licensed to be used indefinitely. This is an important part of the collection process, since a majority of current cross-view datasets~\cite{cvusa,vo2016localizing,cvact, vigor,university1652,cvcities,ye2024cross} use data that is either restricted in redistribution or usage for machine learning purposes~\cite{helbich2024use}. The ground view images are licensed under Creative Commons or Public Domain licenses, while the aerial images from the NAIP are considered public domain information. 

\section{Training Multimodal Models}
\label{sec:method}

We employ a simple yet effective baseline model to learn a shared embedding space across all modalities (i.e., ground-view and aerial-view images, text, and GPS coordinates) of \dname. Our approach enables cross-modal retrieval via nearest neighbor search in the joint space.

\mypar{Architecture}
A dedicated encoder processes each modality. We use a frozen CLIP image encoder for both ground and aerial images, the corresponding frozen text encoder for Wikipedia text, and initialize a location encoder following GeoCLIP~\cite{geoclip} for GPS coordinates. 
After each encoder, we add a projection head with two trainable linear layers separated by a \texttt{ReLU} activation, following~\cite{geoclip}. This allows projecting modality-specific representations to a common dimensionality. 

\mypar{Loss Function}
To align the modalities, we extend the InfoNCE~\cite{oord2018representation} loss to all pairwise combinations of the four modalities. Given a batch of matching samples across $K=4$ modalities, the loss is defined as:

\begin{equation}
\mathcal{L} = \frac{1}{K(K-1)} \sum_{i=1}^K \sum_{\substack{j=1 \ j \ne i}}^K \mathcal{L}_{i,j}
\label{eq:complete_loss}
\end{equation}

\noindent where $\mathcal{L}_{i,j}$ is the contrastive loss between embeddings from modality $i$ and $j$. During inference, retrieval is performed using $k$-nearest neighbors in the learned embedding space.

\begin{figure*}[t]
    \centering
    \includegraphics[width=0.9\textwidth]{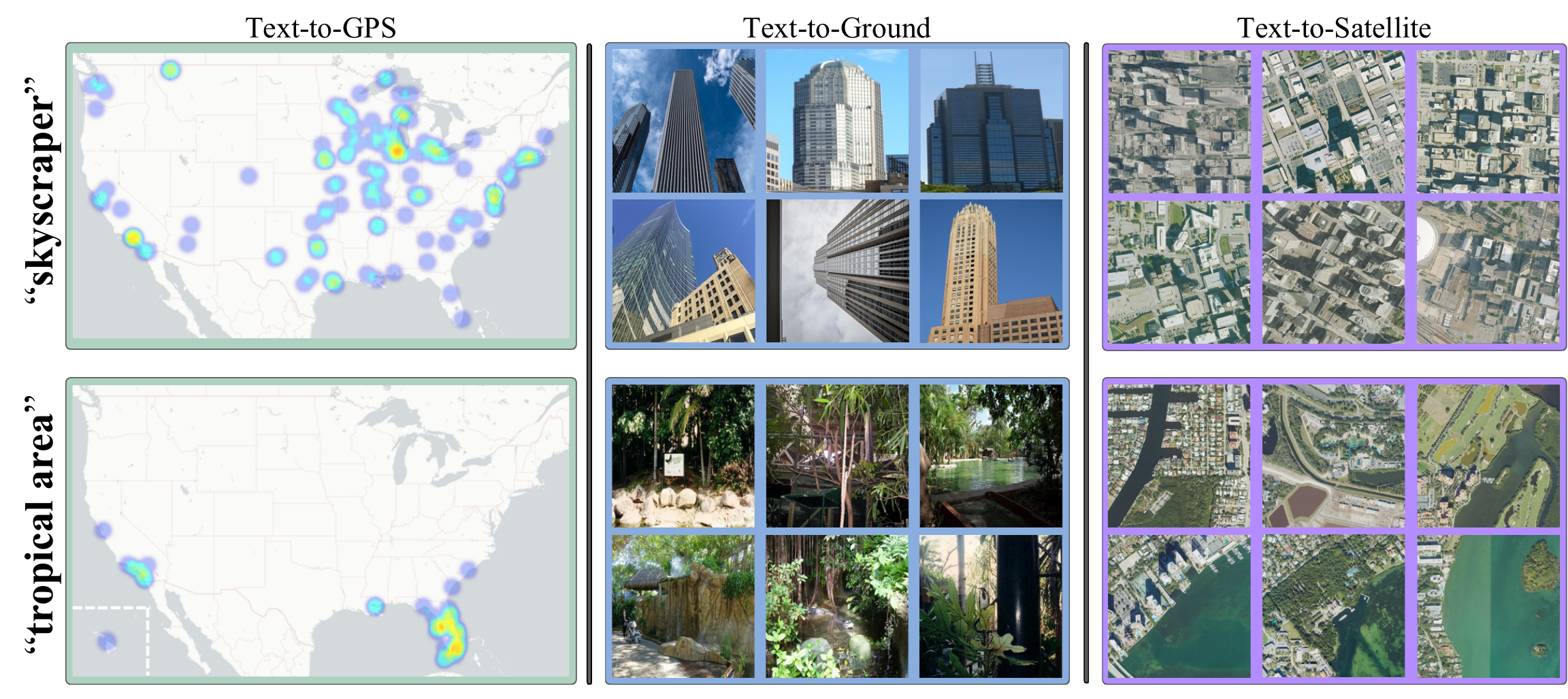}
    \caption{Text-to-GPS (top $1000$), Text-to-Ground and Text-to-Satellite retrieval from the index set with the baseline model. The model accurately locates regions and images that are semantically relevant to the prompt, illustrating strong feature alignment across modalities.}
    \label{fig:visualization}
\vspace{-3mm}
\end{figure*}

\mypar{Implementation details}
All models are trained on a single NVIDIA H100 GPU for $20$ epochs with an AdamW~\cite{loshchilov2017decoupled} optimizer, a learning rate of $1 \times 10^{-4}$, a weight decay of $5\times10^{-4}$, a batch size of $512$, and projection layers of dimension $512$. During training, we randomly select one ground and aerial image for each landmark in the minibatch. We experiment with sampling aerial images randomly versus always using the latest aerial image available.
As a simple strategy, we use sentences from each landmark's text description. We both sample sentences randomly and use the first sentence during training. All results are compared in Tab.~\ref{tab:all_model_experiments}.
To avoid overfitting, we use a validation set with $1024$ landmarks sampled and removed from \dname's training set, and the best model is chosen according to the lowest validation loss. 
We train with our complete pairwise loss across all modalities and following~\cite{geoclip,radford2021}, we set the scaling temperature to $0.07$.
\section{Experiments}
\label{sec:experiments}

In this section, we evaluate the baseline trained on \dname and several off-the-shelf state-of-the-art models on various geo-spatial tasks:
image-to-image retrieval (Sec.~\ref{sec:experiments_i2i}),
ground and satellite geolocalization (Sec.~\ref{sec:experiments_geo}), and text-to-$X$ retrieval (Sec.~\ref{sec:experiments_t2a}).
In Sec.~\ref{sec:experiments_ablation}, we analyze design choices of our baseline model.
The comparison between the \textbf{\textit{Zero-Shot}} performance of off-the-shelf models and our model trained on \dname aims to illustrate: 1) that current off-the-shelf foundation models adapt poorly to geo-spatial tasks, and 2) that specialized models are limited to only solving one specific task. We train \textbf{MMCLIP} as a baseline whose results reflect the potential of training models with multimodal geo-spatial data, yielding satisfactory results across all tasks. Its performance is not meant to claim superiority over any model.

\subsection{Image-to-Image retrieval}
\label{sec:experiments_i2i}

We analyse the results for cross-view ground and satellite retrieval (\cref{tab:ground_sat_table}). 
The capabilities of DINO(v2/v3), CLIP, and SigLIP(2) are evaluated off-the-shelf, and we test the zero-shot performances of Uni-1652, TransGeo-$90^{\circ}$ FoV, and Sample4Geo-UNI, in-domain models that align with our data modalities. Models relying on panoramic ground view images are omitted due to the significant domain gap.

\begin{table}[t]
    \caption{\textbf{Ground-to-GPS geolocalization.} Performance on the \dname query set ($18{,}688$ ground images). Ground-to-Sat-to-GPS is also done to evaluate cross-view retrieval models.}
    \label{tab:ground_gps_table}
    \centering
    \resizebox{0.46\textwidth}{!}{
        \begin{tabular}{ccccccc}
            \toprule
            & \multirow{3}{*}{\makecell{\textbf{Method}}} & \multicolumn{5}{c}{\makecell{Distance (\textbf{\% @ km}) $\uparrow$}} \\
            & & Street & City & Region & Country & Continent\\
            & & (\textbf{1 km}) & (\textbf{25 km}) & (\textbf{200 km}) & (\textbf{750 km}) & (\textbf{2500 km}) \\
            \midrule
            & & \multicolumn{5}{c}{\makecell{Ground $\rightarrow$ Sat $\rightarrow$ GPS}} \\
            \multirow{6}{*}{\rotatebox{90}{\textbf{\textit{Zero-Shot}}}} 
            & OAI-CLIP~\cite{radford2021}(ViT-B/224) & 4.06 & 13.08 & 19.95 & 43.68 & 81.18 \\
            & OAI-CLIP~\cite{radford2021}(ViT-L/336) & 8.04 & 23.53 & 35.50 & 62.41 & 86.77 \\
            & SigLIP2~\cite{siglip2}(ViT-L/512)  & 12.63 & 24.63 & 34.44 & 60.86 & 89.86\\
            & Uni-1652~\cite{university1652} & 0.02 & 1.15 & 5.69 & 22.57 & 65.52 \\
            & Sample4Geo-UNI~\cite{sample4geo} & 0.27 & 2.05 & 6.86 & 25.42 & 66.22 \\
            & TransGeo-$90^{\circ}$~\cite{zhu2022transgeo} & 0.28 & 3.87 & 9.52 & 31.30 & 69.73 \\
            \midrule
            \rowcolor{lightgray}
            & \textbf{MMCLIP} & \underline{18.41} & \textbf{37.40} & \underline{50.99} & \underline{73.72} & \textbf{91.83} \\ 
            \midrule
            \midrule
            & & \multicolumn{5}{c}{\makecell{Ground $\rightarrow$ GPS}} \\
            \multirow{6}{*}{\rotatebox{90}{\textbf{\textit{Zero-Shot}}}} 
            & Always NYC & 0.00 & 17.73 & 22.31 & 47.07 & 74.86 \\
            & GeoReasoner~\cite{li2024georeasoner} & 0.01 & 5.22 & 8.03 & 25.83 & 80.28 \\
            & StreetCLIP~\cite{haas2023learning} & 0.53 & 33.48 & 48.89 & 71.29 & 88.48 \\
            & osv5m~\cite{astruc2024osv} & 1.28 & 13.44 & 19.19 & 37.50 & 57.25 \\
            & G3~\cite{jia2024g3} & 12.86 & 30.46 & 45.52 & 69.36 & \textbf{91.83} \\
            & GeoCLIP~\cite{geoclip} & \textbf{21.37} & \underline{36.44} & 48.57 & 71.45 & 91.50 \\
            \midrule
            \rowcolor{lightgray}
            & \textbf{MMCLIP} & 16.83 & 35.95 & \textbf{51.78} & \textbf{74.94} & \underline{91.52} \\ 
            \bottomrule
        \end{tabular}
        }
\vspace{-2mm}
\end{table}

\noindent \textbf{Metrics.}
We adopt the standard Recall@$K$ (R@$K$) metric, median rank (medR), and mean Average Precision@$1$k (mAP@$1$k) for image-to-image retrieval. The Recall values show how often a correct image is retrieved in the $K^{th}$ position from the index set, while the median rank indicates the median position at which the first correct match is retrieved. The mAP@$1$k is a variant of the mAP metric where only the top-1000 ranked images are considered~\cite{gldv2}.


\noindent \textbf{Results.}
All foundation models improve with larger architectures and higher training resolution for both tasks, with SigLIP2 yielding the most promising results. On the other hand, the specialized models perform poorly despite being trained on cross-view benchmarks. Their results illustrate the lack of diversity in previous cross-view datasets, also highlighted by~\cite{sample4geo}.
Our \textbf{MMCLIP} baseline achieves good performance on both retrieval tasks, but is still far from saturating the benchmark.
%


\subsection{Geolocalization}
\label{sec:experiments_geo}

Geolocalization performance is evaluated with specialized models trained exclusively for this task.
Additionally, we explore a novel setting and assume that a given satellite index set also includes information about the images' GPS locations. We can now also evaluate cross-view models indirectly for this task, taking the GPS coordinates associated with the first retrieved satellite image as the prediction. 
We report the Ground-to-GPS, both directly and indirectly (Tab.~\ref{tab:ground_gps_table}), and Satellite-to-GPS geolocalization (Tab.~\ref{tab:satellite_gps_table}). 

\noindent \mypar{Metrics.}
We follow prior work~\cite{haas2024pigeon, geoclip} and use the distance percentage at different kilometer thresholds (Distance(\% @ km)). This is the percentage of predictions that fall within a specific distance from the true GPS coordinate. We define the GPS index set as the coordinates from the combined satellite index and query sets, yielding a gallery of $101k$ GPS locations. The Haversine distance is used to evaluate the angular distances.

\begin{table}[t]
    \caption{\textbf{Satellite-to-GPS geolocalization.} Model performance on the \dname query set ($1000$ satellite images).}
    \label{tab:satellite_gps_table}
    \centering
    \resizebox{0.46\textwidth}{!}{
        \begin{tabular}{ccccccc}
            \toprule
            & \multirow{3}{*}{\makecell{Satellite $\rightarrow$ GPS\\ \\ \textbf{Method}}} & \multicolumn{5}{c}{\makecell{Distance (\textbf{\% @ km}) $\uparrow$}} \\
            & & Street & City & Region & Country & Continent\\
            & & (\textbf{1 km}) & (\textbf{25 km}) & (\textbf{200 km}) & (\textbf{750 km}) & (\textbf{2500 km}) \\
            \midrule
            \multirow{7}{*}{\rotatebox{90}{\textbf{\textit{Zero-Shot}}}} 
            & SatCLIP~\cite{satclip} & 0.0 & 0.0 & 0.3 & 5.1 & 50.0 \\
            & Always NYC & 0.0 & 8.8 & 15.4 & 41.7 & 74.2 \\
            & GeoReasoner~\cite{li2024georeasoner} & 0.6 & 3.1 & 6.2 & 24.1 & 82.9 \\
            & StreetCLIP~\cite{haas2023learning} & 0.6 & 27.2 & 49.6 & 84.5 & 93.9 \\
            & osv5m~\cite{astruc2024osv} & 0.7 & 11.1 & 21.2 & 44.0 & 70.8 \\
            & G3~\cite{jia2024g3} & 8.8 & 26.9 & 47.9 & 79.8 & 97.0 \\
            & GeoCLIP~\cite{geoclip} & \underline{12.3} & \underline{31.3} & \underline{48.8} & \underline{81.3} & \underline{97.4} \\
            \midrule
            \rowcolor{lightgray}
            & \textbf{MMCLIP} & \textbf{36.9} & \textbf{61.5} & \textbf{81.1} & \textbf{95.5} & \textbf{99.7} \\ 
            \bottomrule
        \end{tabular}
    }
\vspace{-4mm}
\end{table}

\noindent \mypar{Results.}
As a baseline, we take the geographic center of the most densely sampled location in \dname, New York City, and evaluate predicting that location for all queries ("Always NYC").
CLIP~\cite{radford2021} and SigLIP~\cite{siglip2} perform considerably well for indirect geolocalization.
%
The poor performance of the models specialized in cross-view localization is expected, since they are trained on data from limited regions, inherently with low image diversity.
GeoReasoner and StreetCLIP predict city names, which we adapt to estimate the geolocation at the city's center. Consequently, their performance deteriorates for fine-grained predictions.
osv5m has similar results to those reported in the paper.
GeoCLIP and G3 show impressive zero-shot performance, outperforming \textbf{MMCLIP} respectively on the street-level and continent-level.
With significantly fewer training images than task-specific models like GeoCLIP and G3, our baseline still achieves comparable results on most geolocalization metrics. Because \dname focuses on well-known landmarks, it is likely that some ground images from the query and index sets also appear in MP16~\cite{mp16}, which can inflate the performance of GeoCLIP and G3. 
%
%
%
In the Satellite-to-GPS task, SatCLIP adapts poorly to high-resolution imagery because of the large domain gap between the Sen-2 and NAIP images.
G3 and GeoCLIP achieve competitive results, reflecting a semantic understanding of the query images, albeit with less precision on fine-grained street-level details. 
\textbf{MMCLIP} is able to correctly locate a large proportion of the query images.

\begin{table}[t]
    \caption{\textbf{Text-to-Any retrieval.} Model performance on the \dname query set ($1000$ first sentences) for Text-to-Satellite and Text-to-GPS retrieval.}
    \label{tab:text_to_image_table}
    \centering
    \resizebox{0.46\textwidth}{!}{
        \begin{tabular}{ccccccc}
            \toprule
            & \textbf{Retrieval} & medR $\downarrow$ &  mAP@1k$\uparrow$  & R@1 $\uparrow$ & R@5 $\uparrow$ & R@10 $\uparrow$ \\
            \midrule
            & & \multicolumn{5}{c}{\makecell{Text $\rightarrow$ Satellite}} \\
            \multirow{3}{*}{\rotatebox{90}{\textbf{\textit{Z-Shot}}}} 
            & OAI-CLIP (ViT-B/224)          & 2449 & 8.1 & 5.7 & 15.7 & 21.6 \\
            & OAI-CLIP (ViT-L/336)          & \underline{1037} & \underline{14.5} & \underline{11.1} & \underline{25.2} & \underline{33.3} \\
            & SigLIP2 (ViT-L/512)           & 3295 & 7.16 & 4.9 & 14.0 & 18.4 \\
            \rowcolor{lightgray}
            \midrule
            & \textbf{MMCLIP}               & \textbf{388} & \textbf{17.3} & \textbf{13.4} & \textbf{31.0} & \textbf{41.4} \\ 
            \midrule
            \midrule
            & \multirow{2}{*}{\makecell{\textbf{Localization}}} & Street$\uparrow$ & City$\uparrow$ & Region$\uparrow$ & Country$\uparrow$ & Continent$\uparrow$\\
            & & (\textbf{1 km}) & (\textbf{25 km}) & (\textbf{200 km}) & (\textbf{750 km}) & (\textbf{2500 km}) \\
            \midrule
            & & \multicolumn{5}{c}{\makecell{Text $\rightarrow$ GPS}} \\
            \rowcolor{lightgray}
            & \textbf{MMCLIP} & \textbf{4.2} & \textbf{16.5} & \textbf{29.9} & \textbf{55.2} & \textbf{86.1} \\ 
            \bottomrule
        \end{tabular}
    }
\vspace{-4mm}
\end{table}

\subsection{Text-to-Any Retrieval}
\label{sec:experiments_t2a}
We evaluate the baseline model on Text-to-Any retrieval in Tab.~\ref{tab:text_to_image_table}. Text retrieval and geolocation are performed by encoding the first sentence of each query Wikipedia text and finding the most similar images/locations. The first sentences in Wikipedia often include strong geographical cues, which may inflate performance. We remove location cues with GPT 3.5~\cite{brown2020language}, and manually adjust the modified sentences to ensure that no explicit place names are mentioned.

\noindent \textbf{Metrics.} The same metrics are used as in  \cref{sec:experiments_i2i} and \ref{sec:experiments_geo}. For Text-to-Satellite, only one positive match is available per query in the index set, while for Text-to-GPS, the haversine distance is computed between the query's ground truth and the first retrieved GPS location from the $101k$ gallery.

\noindent \textbf{Results.} Text-to-Satellite and Text-to-GPS results reflect considerable geospatial understanding, but fall short compared to results in Tab. \ref{tab:ground_sat_table} and \ref{tab:ground_gps_table}. The query sentences contain only vague semantic information compared to the visual queues in cross-view retrieval.
For a better understanding of the textual alignment with the other modalities, we show examples in \cref{fig:visualization} of Text-to-$X$ retrieval given a simple text prompt. We present the first $1000$ nearest GPS coordinates, the most similar retrieved ground and satellite images, reflecting a clear correspondence in the joint space.

\setlength{\tabcolsep}{2pt}
\begin{table}[t]
    \caption{\textbf{Ablation studies.} Performance for different models trained in various configurations: training objectives, modalities included, text sampling (F: first sentence or R: random sentence), satellite sampling (R: random or L: last), and whether indoor ground images are included during training (Subset). The gray row is our final baseline model, used in all tables.}
    \label{tab:all_model_experiments}
    \centering
    \resizebox{0.47\textwidth}{!}{
        \begin{tabular}{ccccc|ccc}
            \toprule
            Objective & Modalities & Text & Sat & Subset & mAP@1k$_{S \rightarrow G}$ & mAP@1k$_{G \rightarrow S}$ & G/S$\rightarrow$GPS (\textbf{1 km}) \\
            \midrule
            \multirow{7}{*}{\makecell{all$\Leftrightarrow$all}}
            & G,S & \xmark & R & \xmark    & 17.59 & 25.59 & \xmark \\ 
            & G,S,T & F & R & \xmark       & 17.31 & 25.61 & \xmark \\ 
            & G,S,T & R & R & \xmark       & 17.29 & 26.16 & \xmark \\ 
            & G,S,C & \xmark & R & \xmark  & 16.19 & 22.96 & 16.67 / \underline{27.9} \\ 
            & G,S,T,C & F & R & \xmark     & 17.39 & 25.05 & 15.63 / 27.7 \\ 
            & G,S,T,C & R & R & \xmark     & 16.93 & 24.94 & 16.27 / 25.4 \\ 
            & G,S,T,C & R & R & \cmark     & 17.45 & 25.07 & \textbf{17.71} / 27.2 \\ 
            \rowcolor{lightgray}
            \textbf{MMCLIP} & \cellcolor{lightgray}G,S,T,C & \cellcolor{lightgray}R & \cellcolor{lightgray}L & \cellcolor{lightgray}\cmark & \cellcolor{lightgray}\underline{18.79} & \cellcolor{lightgray}\underline{26.20} & \cellcolor{lightgray}\underline{16.83} / \textbf{36.9} \\ 
            \midrule
            \multirow{3}{*}{\makecell{G $\Leftrightarrow$ all}}
            & G,S,T,C & R & R & \xmark     & 16.5 & 24.86 & 16.55 / 15.7 \\ 
            & G,S,T,C & R & R & \cmark     & 17.08 & 25.16 & 16.18 / 15.4 \\ 
            & G,S,T,C & R & L & \cmark     & \textbf{18.89} & \textbf{27.46} & 15.68 / 18.3 \\ 
            \bottomrule
        \end{tabular}
    }
\vspace{-4mm}
\end{table}


\subsection{Ablation studies}
\label{sec:experiments_ablation}

Tab.~\ref{tab:all_model_experiments} shows all experiments performed with different contrastive learning objectives, number of modalities, text and satellite sampling strategies, and data filtration. Increasing the number of modalities appears to slightly decrease retrieval performance.
Random and First sampling offer equivalent results, while taking the most recent satellite image (L: last) and training on the ground view subset with outdoor images significantly improves the baseline across all tasks.
While the best Imagebind (G $\Leftrightarrow$ all) baseline performs slightly better on retrieval tasks, the fully contrastive setup stands out in geolocalization, especially for Satellite-to-GPS, where the Imagebind baseline's features are only indirectly aligned.

\section{Conclusion}
\label{sec:conclusion}

With \dname, we illustrate the necessity for a multimodal geo-spatial dataset by showing that current off-the-shelf specialized and foundation models have limited performance on our benchmark. We show that a baseline trained on our dataset can reach satisfactory results across these multiple tasks, in contrast with previous models that only focus on solving one task.
Geo-spatial learning is of a multimodal nature -- we hope that this benchmark will motivate further research in multimodal learning, cross-view reasoning, and geo-spatial understanding of the world.

\section*{Acknowledgements}

This research was supported by the Infrastructure for Quantitative AI-based Tomography (QUAITOM), under the Novo Nordisk Foundation Data Science Programme grant (Grant number NNF21OC0069766).
Dim P. Papadopoulos was supported by the DFF Sapere Aude Starting Grant ``ACHILLES''.

{
    \small
    \bibliographystyle{ieeenat_fullname}
    \bibliography{main}
}

\clearpage
\maketitlesupplementary

\def\thesection{\Alph{section}}
\setcounter{figure}{3}
\setcounter{table}{6}
\setcounter{section}{0}

\begin{figure*}
    \centering
    \includegraphics[width=\textwidth]{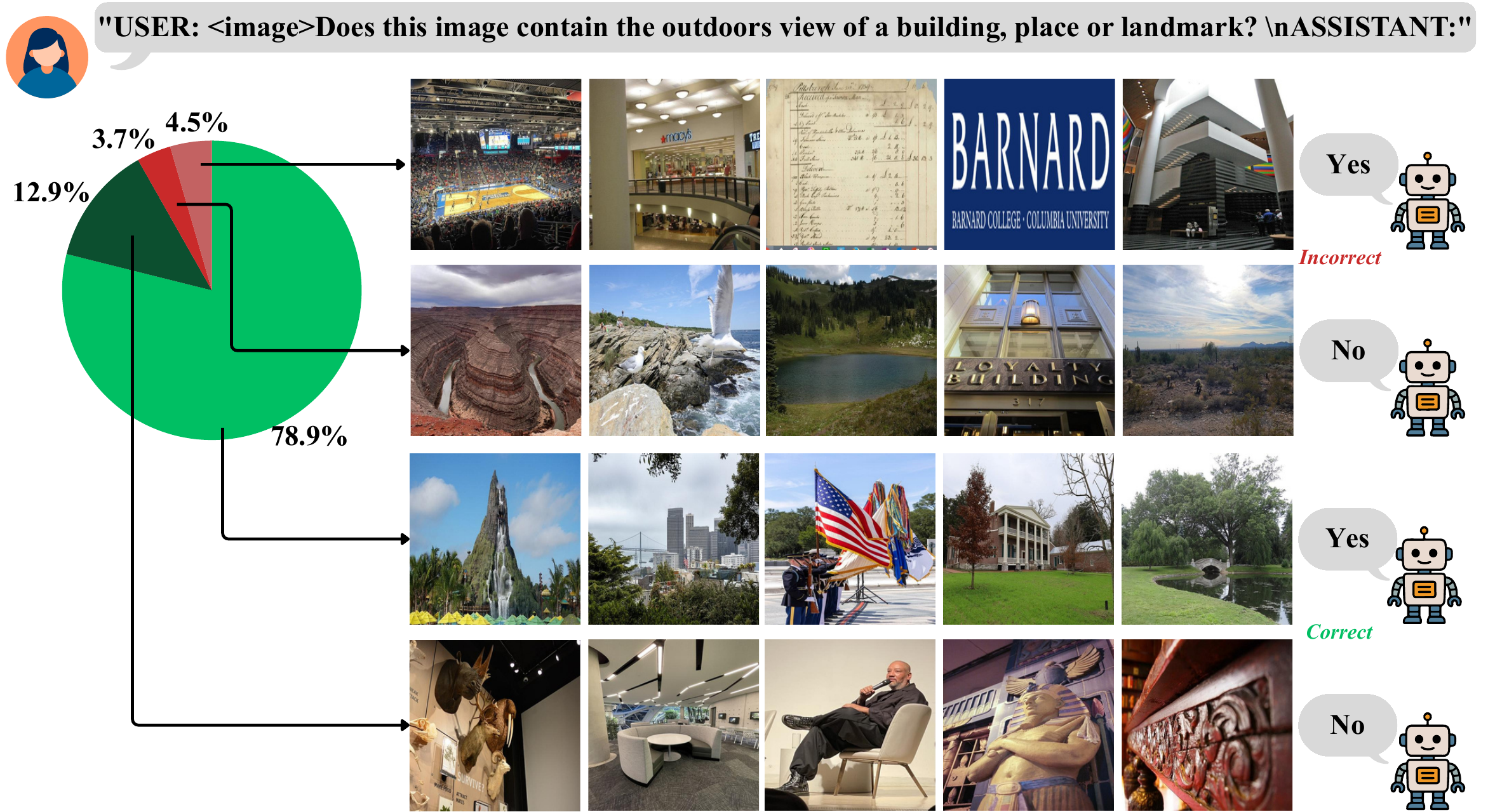}
    \caption{VLM Filtering: Examples of wrong categorizations during the VLM - indoor/outdoor filtering stage. The model may incorrectly classify images as belonging to the opposite category due to visual ambiguities in the images. Five examples are shown from the $8.2\%$ wrongly classified ground images in the $1000$ visually inspected samples.}
    \label{fig:vlm_processing_error}
\end{figure*}

We provide additional information and visualizations for the \dname dataset. Sec.~\ref{appending:data_info} describes the distribution of images per landmark, geographic and categoric distributions, and details the collection pipeline for selecting the landmarks with OpenStreetMaps, Wikimedia Commons and Wikipedia. In Sec.~\ref{appending:visualizations}, more visualizations of the dataset are included, with bounding box examples and a more detailed analysis of each modality illustrated with landmarks from Denver in Figures \ref{fig:mml_denver} and \ref{fig:aerial_denver}. The VLM data processing procedure is mentioned in Sec.~\ref{appendix:vlm_filtering}, with examples of when the VLM fails to correctly categorise the images in Fig.~\ref{fig:vlm_processing_error}. Finally, additional illustrations of landmarks from \dname are presented in Figures \ref{fig:more_examples_1}-\ref{fig:more_examples_4}. 

\begin{algorithm}
\caption{\dname landmark collection pipeline}
\begin{algorithmic}[1]  
\label{algo:collection_pipeline}
\REQUIRE $OSM$ data information, Wikidata, Wikimedia Commons, Wikipedia
\STATE $Dataset = \{\}$
\FOR{each sample $\textbf{polygon}_i$ in $OSM$}
    \STATE $(\text{lat,lon}) = \text{nodes}(\textbf{polygon}_i)$
    \IF{$(\texttt{wikidata}\cup\texttt{wikipedia})\in\textbf{polygon}_i$ tags}
        \STATE Extract Wiki information $Q_i$.
        \STATE ($\text{Commons}_i$ \& $\text{Wikipedia}_i$) $\gets$ \url{wikidata.org/wiki/Qi}
        \IF{$\exists (\text{ground images})\in \text{Commons}_i$}
            \STATE $G_i = True$
        \ENDIF
        \IF{$\exists (\text{wiki text})\in \text{Wikipedia}_i$}
            \STATE $T_i = True$
        \ENDIF
        \IF{$\text{abs}(min(lat)-max(lat)) \cap \text{abs}(min(lon)-max(lon)) < 400m$}
            \STATE $S_i = True$
            \STATE $C_i = True$
        \ENDIF
    \ENDIF
    \STATE $\text{MML}_i = \{G_i,T_i,S_i,C_i\}$
    \IF{$\{M_i=True | \forall M_i \in \text{MML}_i \}$}
        \STATE $Dataset \gets \text{MML}_i$
    \ENDIF
\ENDFOR
\STATE $Dataset$ contains US landmarks with ground, satellite, text, and GPS information.
\end{algorithmic}
\end{algorithm}

\section{Dataset Details}
\label{appending:data_info}

\mypar{Distribution.} As mentioned in Sec.~\ref{sec:dataset}
, the data contains long-tail distributions, both in terms of geographical location, landmark type, and number of images per landmark. In this section, we provide additional information and illustrate these characteristics of the datasets that are inherent in geo-spatial data. 

\begin{figure}
    \centering
    \includegraphics[width=0.40\textwidth]{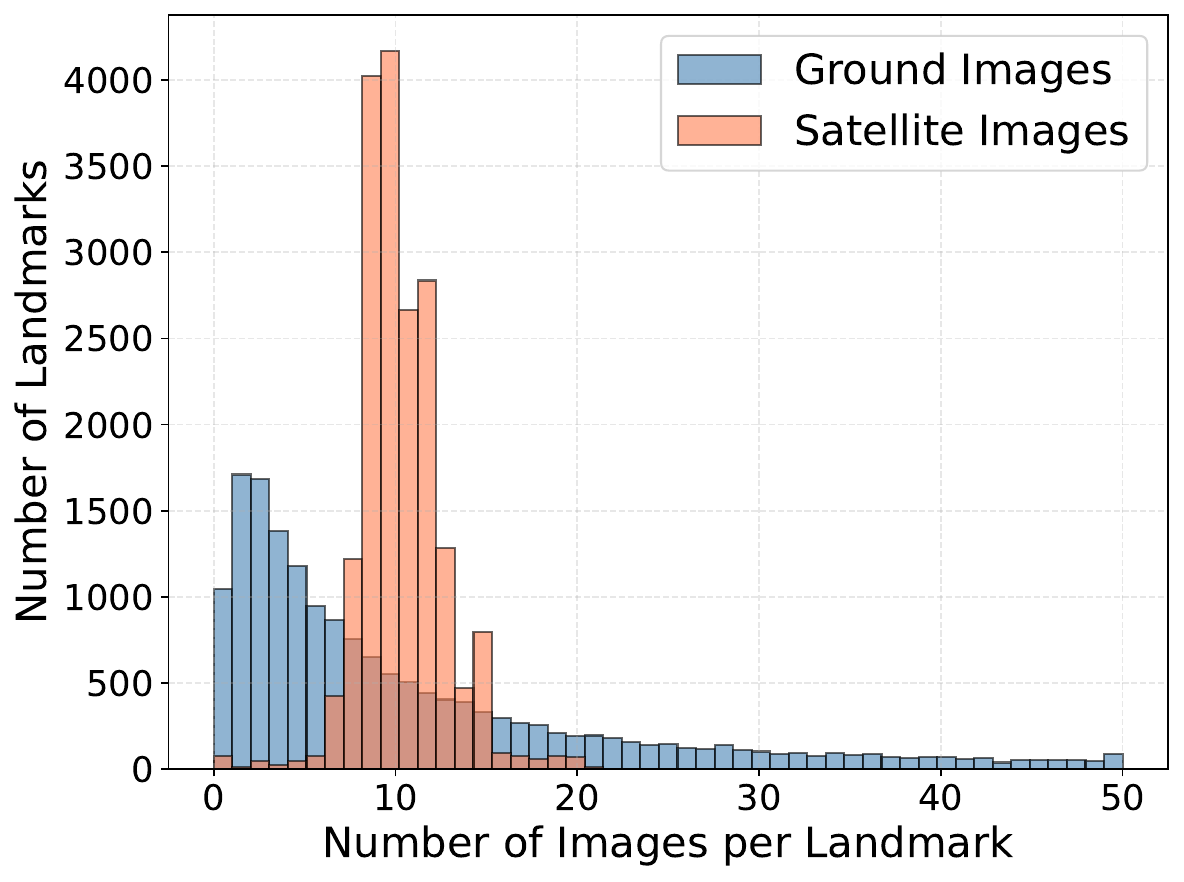}
    \caption{Histogram distribution of the number of images per landmark. A large proportion of landmarks have between 1 and 10 ground images, with a long-tailed distribution. The number of satellite images per landmark follows a bell curve centred at 10 images, with some landmarks having up to 20 aerial images.}
    \label{fig:img_per_landmark}
\end{figure}

In Fig.~\ref{fig:img_per_landmark}, the number of images per landmark is plotted for ground and satellite images. The ground images have a long tail, with most landmarks having 1 to 10 images. The satellite images have a normal distribution, with an average of 10 images per landmark, and some have as many as 20 images, spanning over two decades.

\begin{figure*}
    \centering
    \begin{subfigure}[t]{0.36\textwidth}
        \centering
        \includegraphics[width=\textwidth]{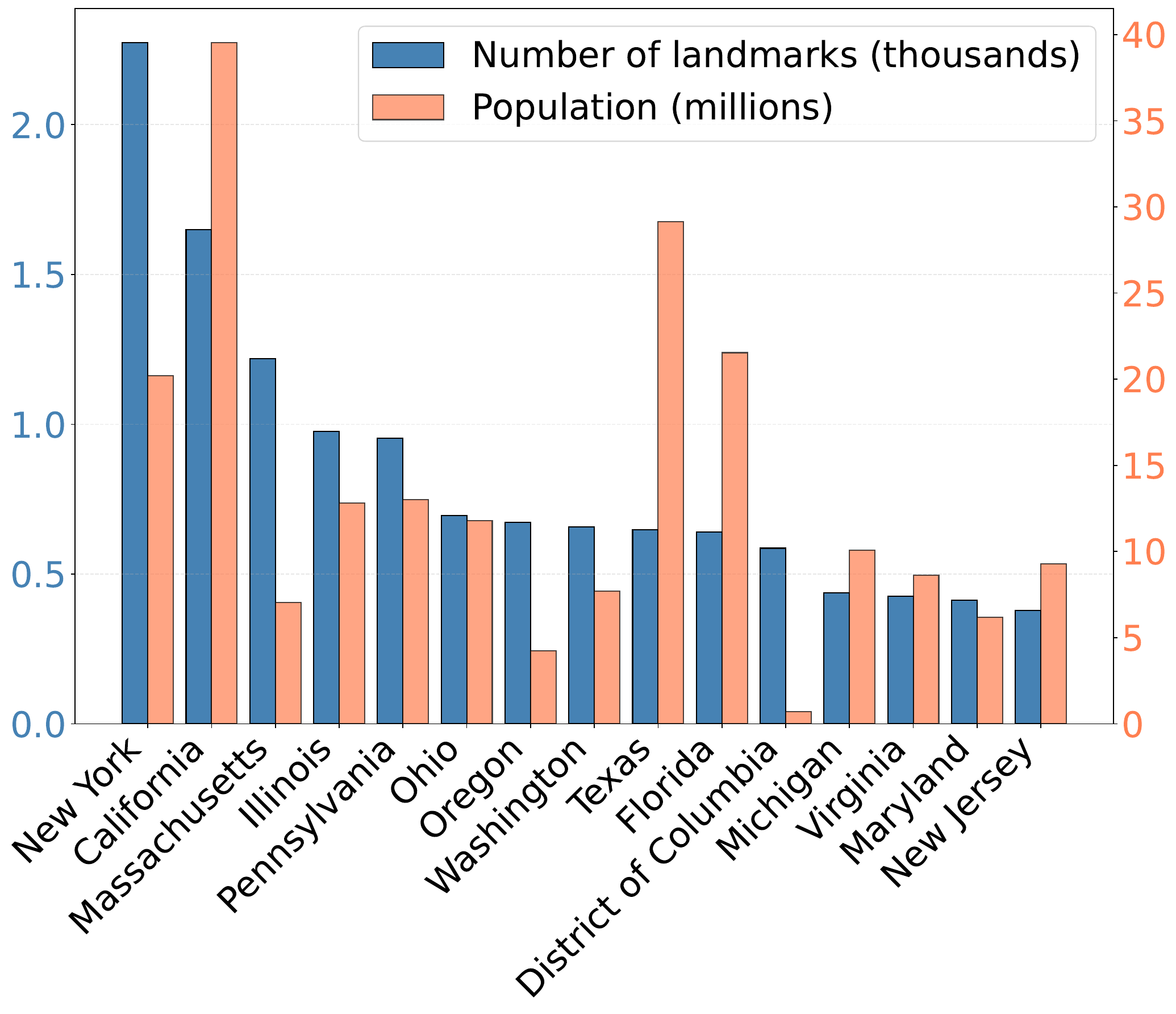}
        \caption{Histogram distribution of the top 15 states with the most landmarks (blue), compared to each state's population (orange). The two distributions diverge, with certain states being disproportionately over-represented or under-represented.}
        \label{fig:land_pop_per_state}
    \end{subfigure}
    \hfill
    \begin{subfigure}[t]{0.56\textwidth}
        \centering
        \includegraphics[width=\textwidth]{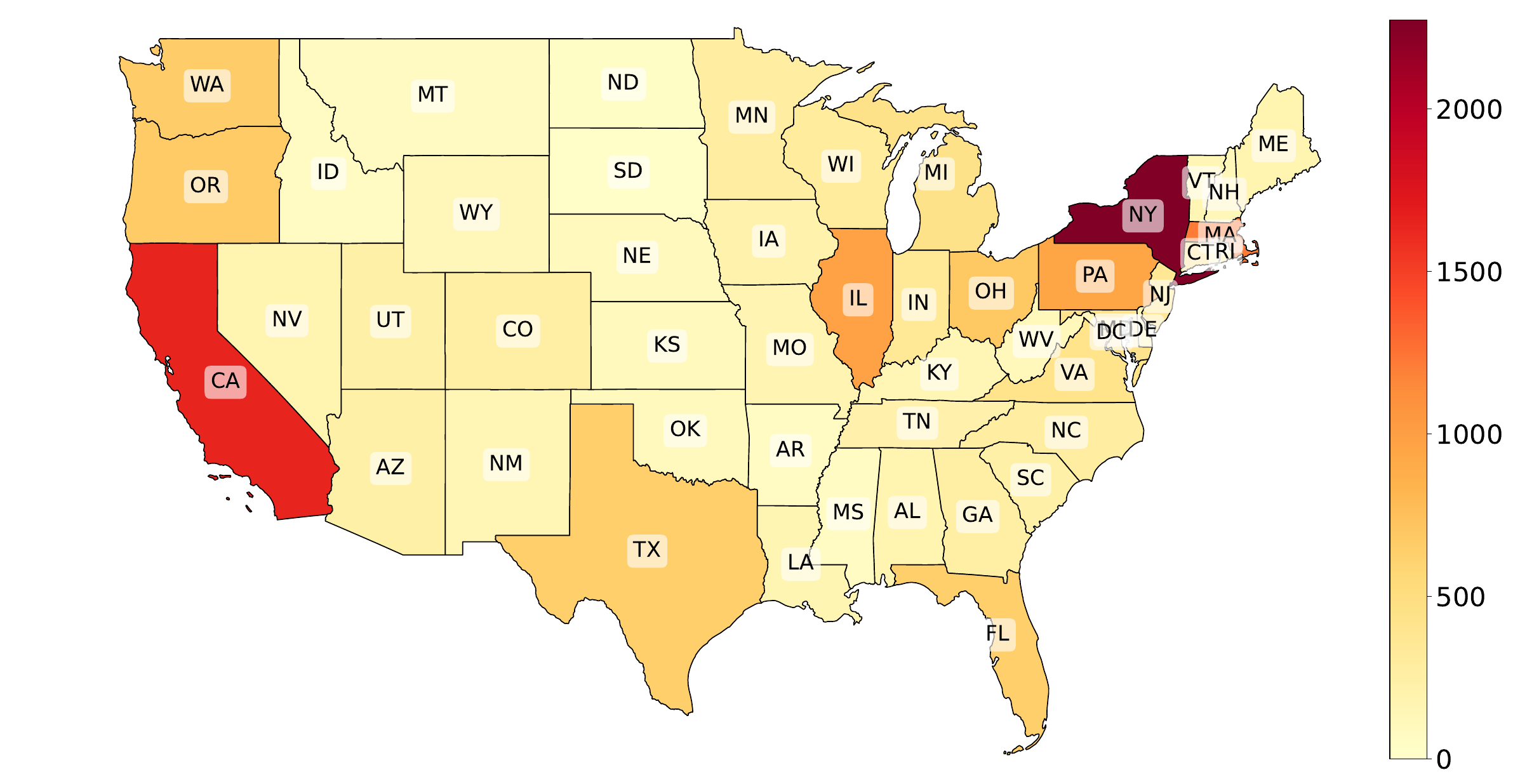}
        \caption{Geographic visualization of the number of landmarks per state in the United States. similar to (a), a large proportion of landmarks come from popular areas such as New York \& California.}
        \label{fig:distribution}
    \end{subfigure}
    \caption{Visual and Geographical illustrations of the landmark distribution across \dname.}
    \label{fig:combined_figures}
\end{figure*}

The histogram in Fig.~\ref{fig:combined_figures} shows the top 15 states with the most landmarks in blue, as well as the population of each state in orange. The number of landmarks does not correlate with the population of the state, with a clear example of the District of Columbia (a federal district which is not part of any state), where many governmental landmarks are located despite a low number of residents. The states with the lowest number of landmarks are Hawaii (22), South Dakota (35), and North Dakota (37).

On the right of Fig.~\ref{fig:combined_figures}, a map of the United States illustrates the geographic distribution of landmarks by state, showing the highest concentrations in states with larger populations, such as New York, California, Texas, and Florida.

\begin{figure*}
    \centering
    \includegraphics[width=\textwidth]{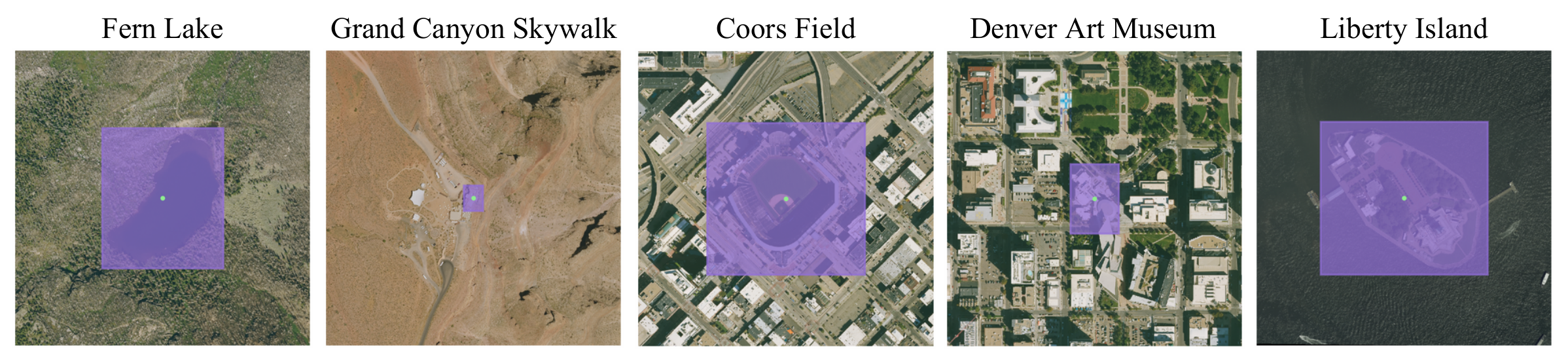}
    \caption{Visualization of the center GPS (green) and bounding boxes (purple) for the polygons associated with different landmarks.}
    \label{fig:bbox_examples}
\end{figure*}

\mypar{Collection pipeline.} In Alg.~\ref{algo:collection_pipeline}, we present our landmark collection pipeline for \dname. For each polygon found from OpenStreetMaps ($OSM$), we filter those that have either a \texttt{wikidata} or \texttt{wikipedia} tag in their description. This grants us the possibility to search for the landmark's Wiki-identifier ($Q_i=\text{Q123456}$), from which the landmark's Wikipedia and Wikimedia Commons webpage can be retrieved. 

If ground images and Wikipedia text are available, we proceed to the final part of the processing pipeline: aerial uniformity. We aim to streamline the collection process from the National Agricultural Imagery Program (NAIP) while ensuring that all images have consistent dimensions. In fact, many landmarks vary in size and shape, sometimes spanning several kilometers (e.g., lakes, railways, mountain ranges, etc.). The scarcity of such large landmarks would cause geospatial ambiguity: to capture the entire landmark, one may need to sample images from a small zoom level, leading to low-resolution imagery. Alternatively, samples from various locations of the landmark at high resolution could be a direction. However, this approach often results in a predominance of uniform images depicting only rocks, water, or small portions of a building.

\begin{table}
    \centering
    \begin{tabular}{l|r}
        Licenses & Counts \\
        \midrule
        CC BY & $70{,}288$ \\
        CC SA & $85$ \\
        CC BY-SA & $181{,}409$ \\
        CC BY-NC-SA & $76$ \\
        Public Domain & $56{,}216$ \\
        CC0 & $16{,}046$ \\
        Attribution & $873$ \\
        No restrictions & $3{,}902$ \\
        \bottomrule
    \end{tabular}
    \caption{Specific licenses for the ground view images in \dname. The licenses and link to each image are also included as part of the dataset for proper attribution.}
    \label{tab:licenses}
\end{table}

This both complicates the pipeline and incurs ambiguity due to the lack of correspondence between the modalities. To this end, we heuristically choose a maximum length of 400 meters for the largest side of the landmark's bounding box. As mentioned in Sec.~\ref{sec:dataset}
, the ground images are resized such that the largest side is 800 pixels. Keeping the same pixel dimensions, we aim to sample aerial images of pixel size $800 \times 800$. After visual inspections, we find that 400 meters yields a balance between the landmark and its surroundings.

\mypar{Licenses.} We collect the \dname dataset, intending to make everything freely accessible and usable. The NAIP aerial imagery is categorized as Public Domain, making it ideal for data collection and incurring no significant filtration in the process. For ground-view images sourced from Wikimedia Commons, an additional step is required to verify the licensing of each image. We keep all the images licensed under ``Creative Commons", ``Public Domain", and other licenses such as ``Attribution" and ``No restrictions" which allow copy, modification, and redistribution. We group the licenses and summarize their count in Tab.~\ref{tab:licenses}.

\begin{figure}
    \centering
    \includegraphics[width=0.40\textwidth]{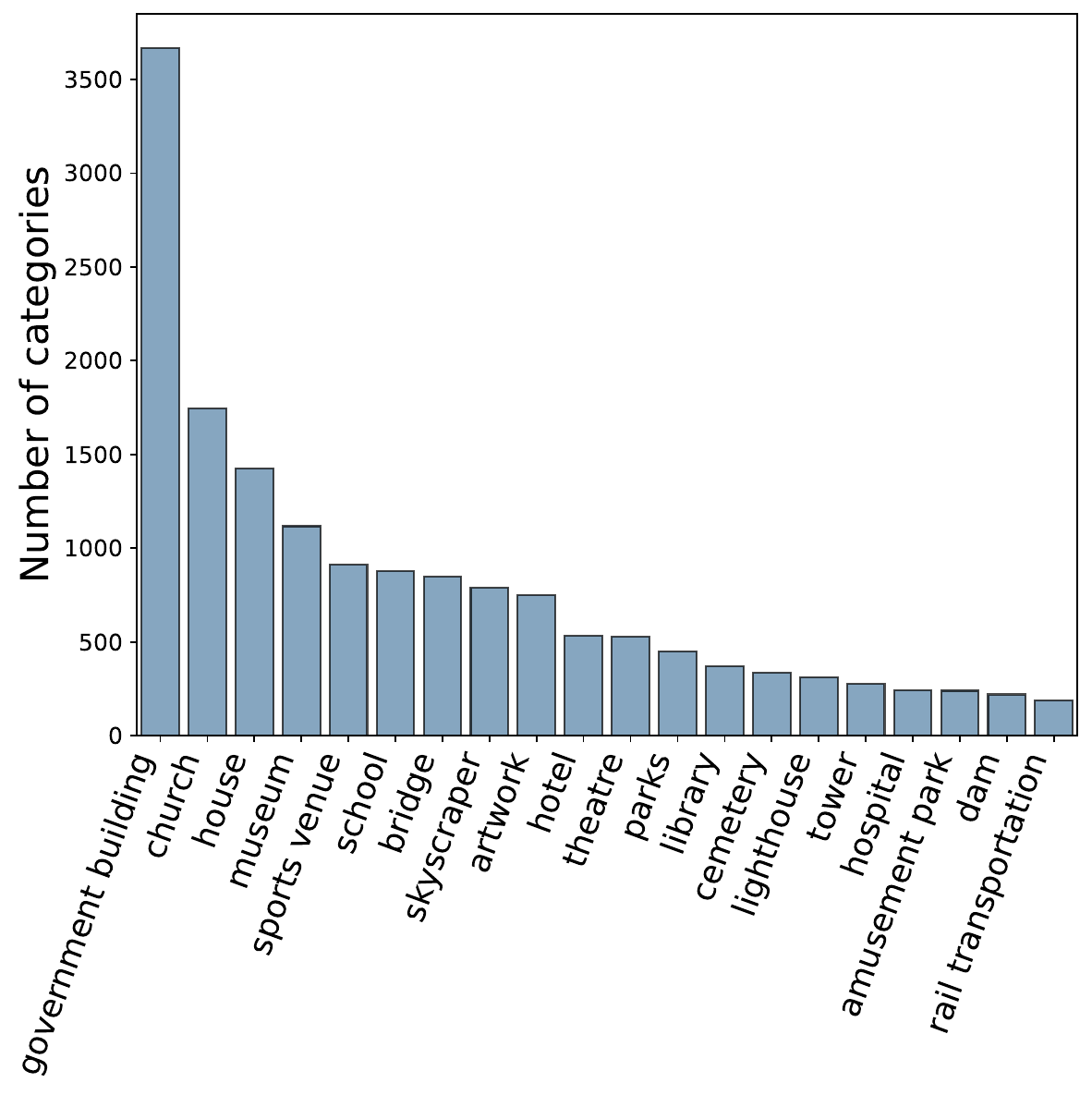}
    \caption{15 most popular super-categories in \dname. Each landmark is categorized as one of the 79 hierarchical super-categories from Ramzi et al.~\cite{HGLDv2}
    . There is a large majority of ``government building" landmarks, which are further analysed in Fig.~\ref{fig:landmarks_per_govment}.}
    \label{fig:landmarks_per_super}
\end{figure}

\begin{figure}
    \centering
    \includegraphics[width=0.40\textwidth]{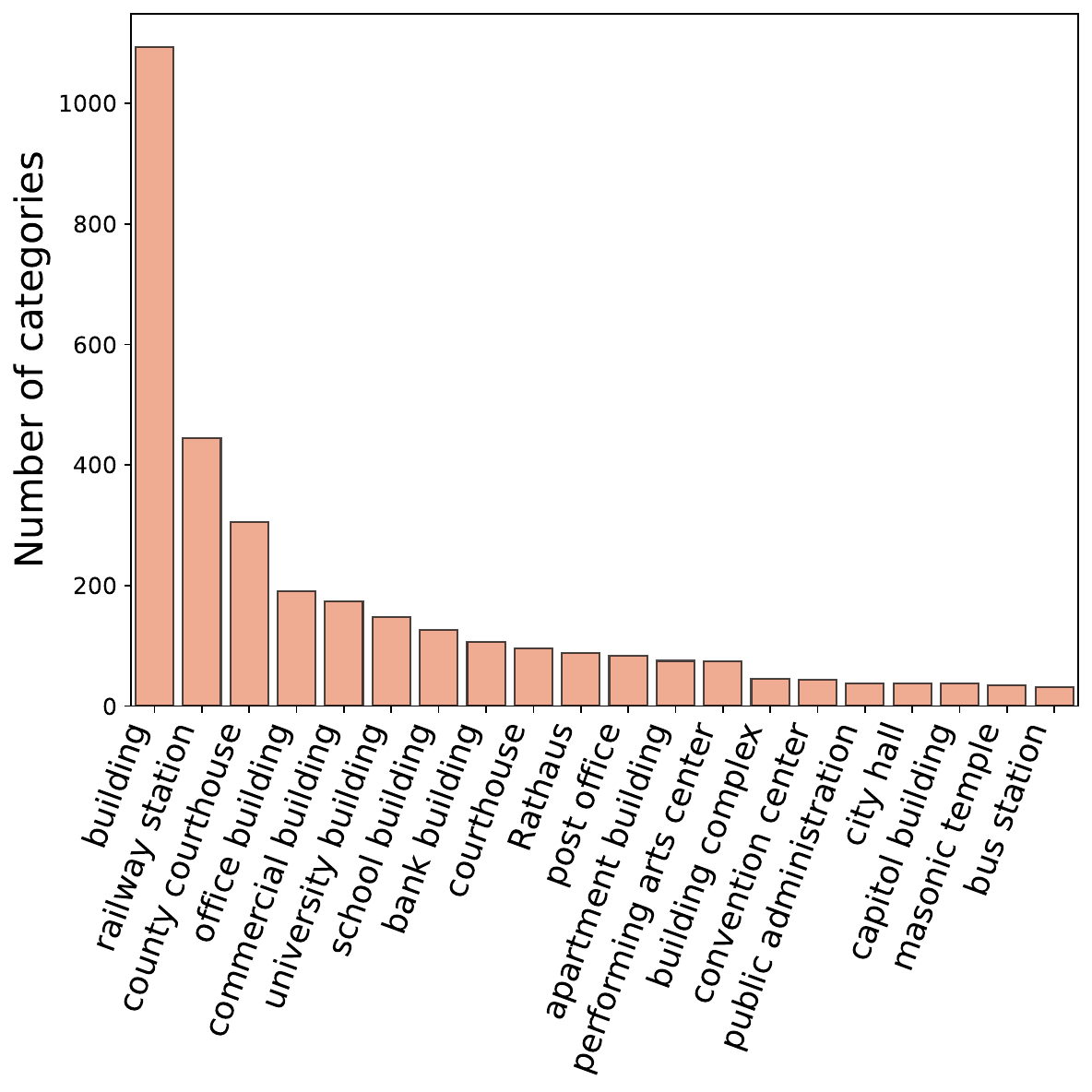}
    \caption{15 most popular categories from the ``government building" super-category. Since a CLIP text encoder is used to classify each category amongst one of the 79 hierarchical categories, all landmark categories with the tag ``building" are placed under ``government building".}
    \label{fig:landmarks_per_govment}
\end{figure}

\section{Additional Visualizations}
\label{appending:visualizations}

\mypar{Super-categories.} As extra information retrieved from each landmark's Wikidata page, we collect tags under the ``Instance of" section for all landmarks as their category, similar to the Hierarchical GLDv2~\cite{HGLDv2}
, in which 79 unique hierarchical super-categories are defined. Once the tags are found, we use a CLIP text encoder to embed each category and assign it to one of the super-categories from Ramzi et al.~\cite{HGLDv2}
. This is achieved by computing the logits between the 79 super-category embeddings and each query category, then selecting the super-category with the highest probability. 
In Fig.~\ref{fig:landmarks_per_super}, we show the number of landmarks per super-category. Because of the large proportion of landmarks identified as ``government building", we also illustrate the sub-categories of the top super-category (see Fig.~\ref{fig:landmarks_per_govment}). We can see that since a lot of landmarks are simply labelled as ``building", when grouped, they are placed under the ``government building" label.

\begin{figure*}
    \centering
    \includegraphics[width=\textwidth]{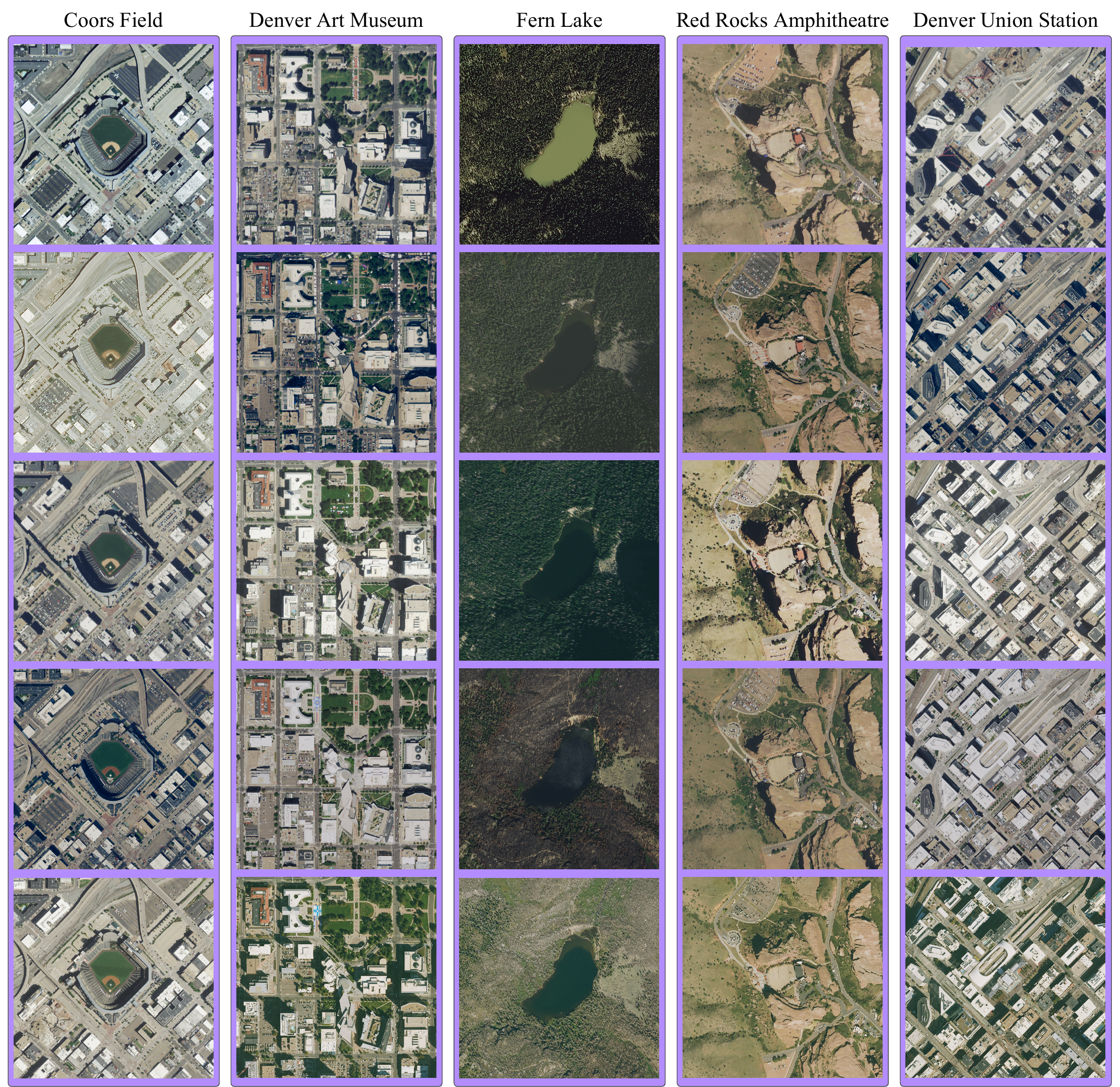}
    \caption{Visual diversity in the aerial imagery from landmarks in \dname. The images are arranged chronologically, ranging from older images at the top to newer images at the bottom. More images of the landmarks are included in the dataset, which are not presented here. The landmarks are sampled from around Denver, Colorado. The temporal aspect of the dataset augments the instance with different moments of capture, angle, and weather conditions. Zoom in to see fine-grained changes in the urban surroundings of columns 1, 2, and 5.}
    \label{fig:aerial_denver}
\end{figure*}

\begin{figure*}
    \centering
    \includegraphics[width=\textwidth]{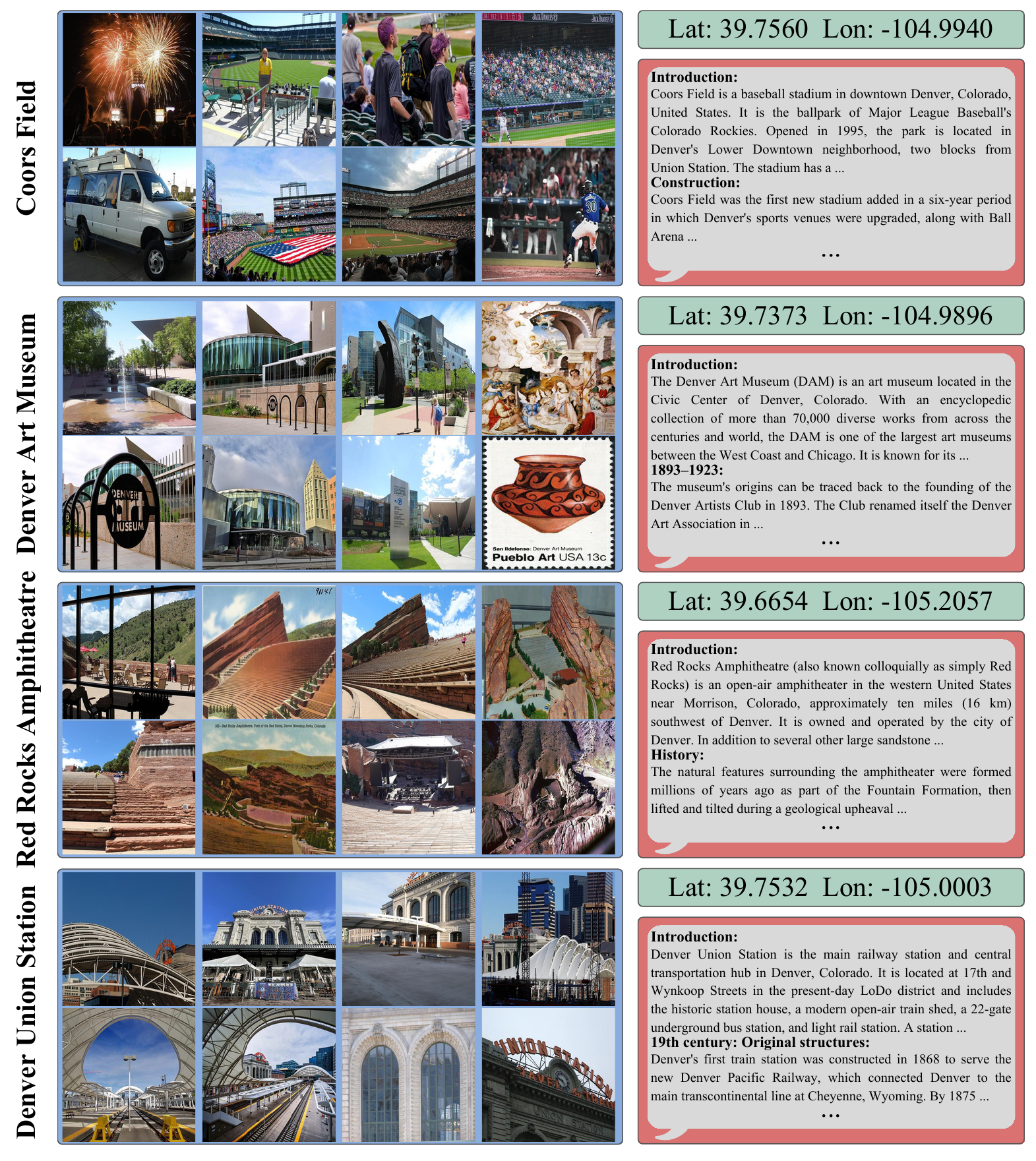}
    \caption{Additional examples of landmarks from \dname. The four landmarks are sampled from around Denver, Colorado.}
    \label{fig:mml_denver}
\end{figure*}

\mypar{More examples.} We show the diversity in \dname by providing examples of landmarks in Fig.~\ref{fig:mml_denver}, as well as temporal changes in the collected aerial imagery in Fig.~\ref{fig:aerial_denver}. We illustrate the intra-state diversity by showing landmarks located around Denver, Colorado. 

Figure~\ref{fig:mml_denver} illustrates the diversity of images associated with each landmark, a characteristic typical of web-sourced imagery. For example, in the case of ``Coors Field", ground-level photos include views captured before the game (image 2), during the game (images 3–7), and after the game (image 1), exhibiting variations in camera angle and lighting conditions. The ``Denver Art Museum" ground views also include scanned documents (image 8) and art from the museum (image 4). Even more intriguing are images 2 and 3 from the Red Rocks Amphitheatre, where the second image is a postcard version from the landmark, from the same angle as the third image. Finally, the last landmark, ``Denver Union Station", reflects the diverse angles and locations from which images are taken of the same landmark, which together give a full understanding of the landmark and its surroundings. Fig.~\ref{fig:mml_denver} also illustrates the geo-spatial fine granularity by providing accurate GPS locations, along with long textual descriptions of the landmark.

Similarly, in Fig.~\ref{fig:aerial_denver}, the aerial images taken from the NAIP illustrate the diversity of the same location when inspecting the same place through time. Not only are all images different in terms of capture time, angle, sun orientation, and season, but physical changes also happen, which can be detected by comparing the images together. Most obvious is the ``Fern Lake", where the size and color of the water vary. It is also evident that tree density was significantly higher in the older (top) images compared to the newer (bottom) views. The fourth image seems to show traces of the wildfires, which have burnt down a majority of the surrounding trees. 

In columns one and four, clear changes appear in the surroundings, with a building appearing in the top left corner of the Coors Field images between images 2 and 3, or a parking space being removed between images 4 and 5.

\begin{figure*}
    \centering
    \includegraphics[width=\textwidth]{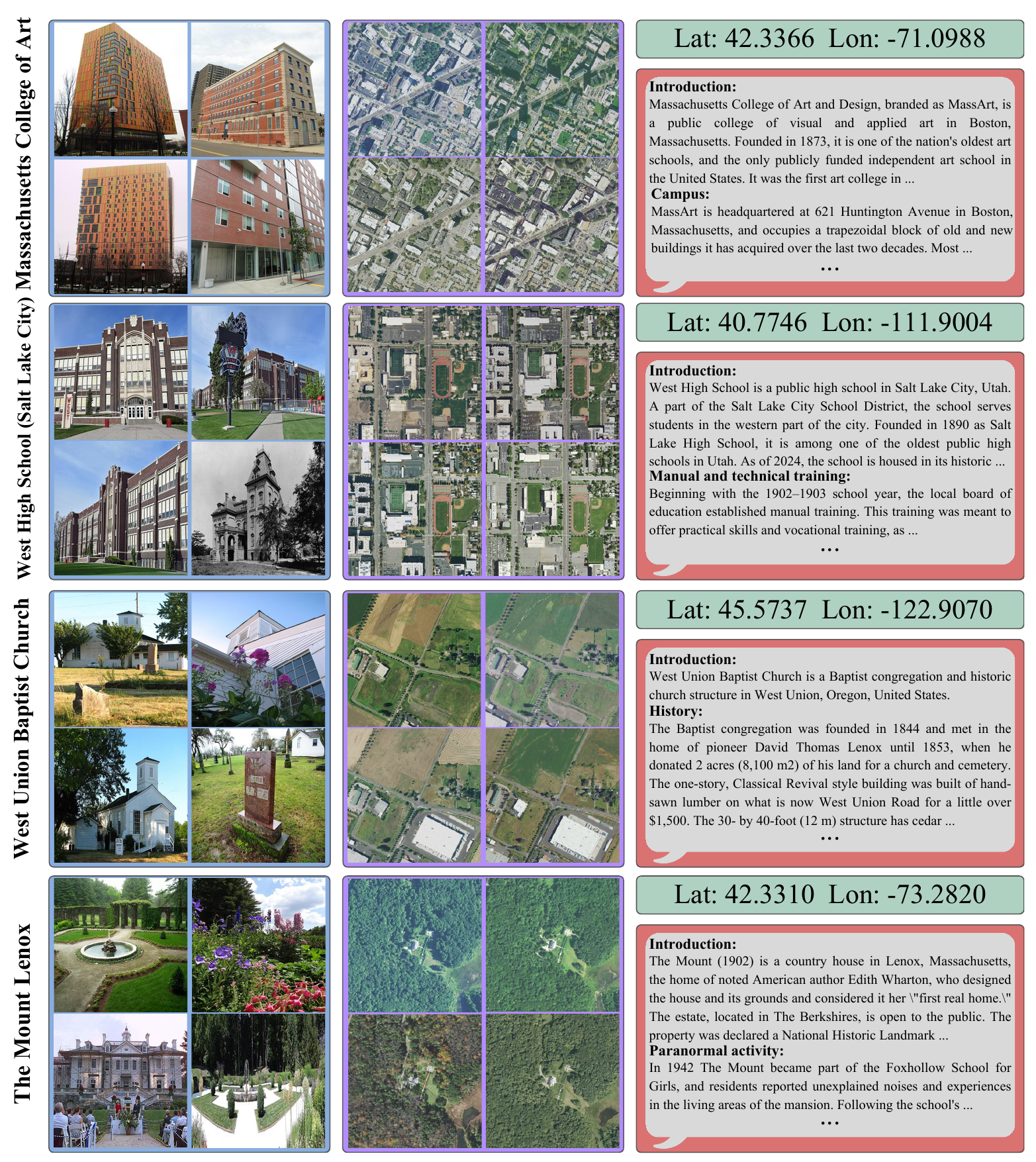}
    \caption{Additional examples from the \dname dataset. We illustrate the diversity in the dataset by randomly sampling landmarks. We show sample ground and satellite views, as well as the exact GPS location and parts of the textual descriptions.}
    \label{fig:more_examples_1}
\end{figure*}

\begin{figure*}
    \centering
    \includegraphics[width=\textwidth]{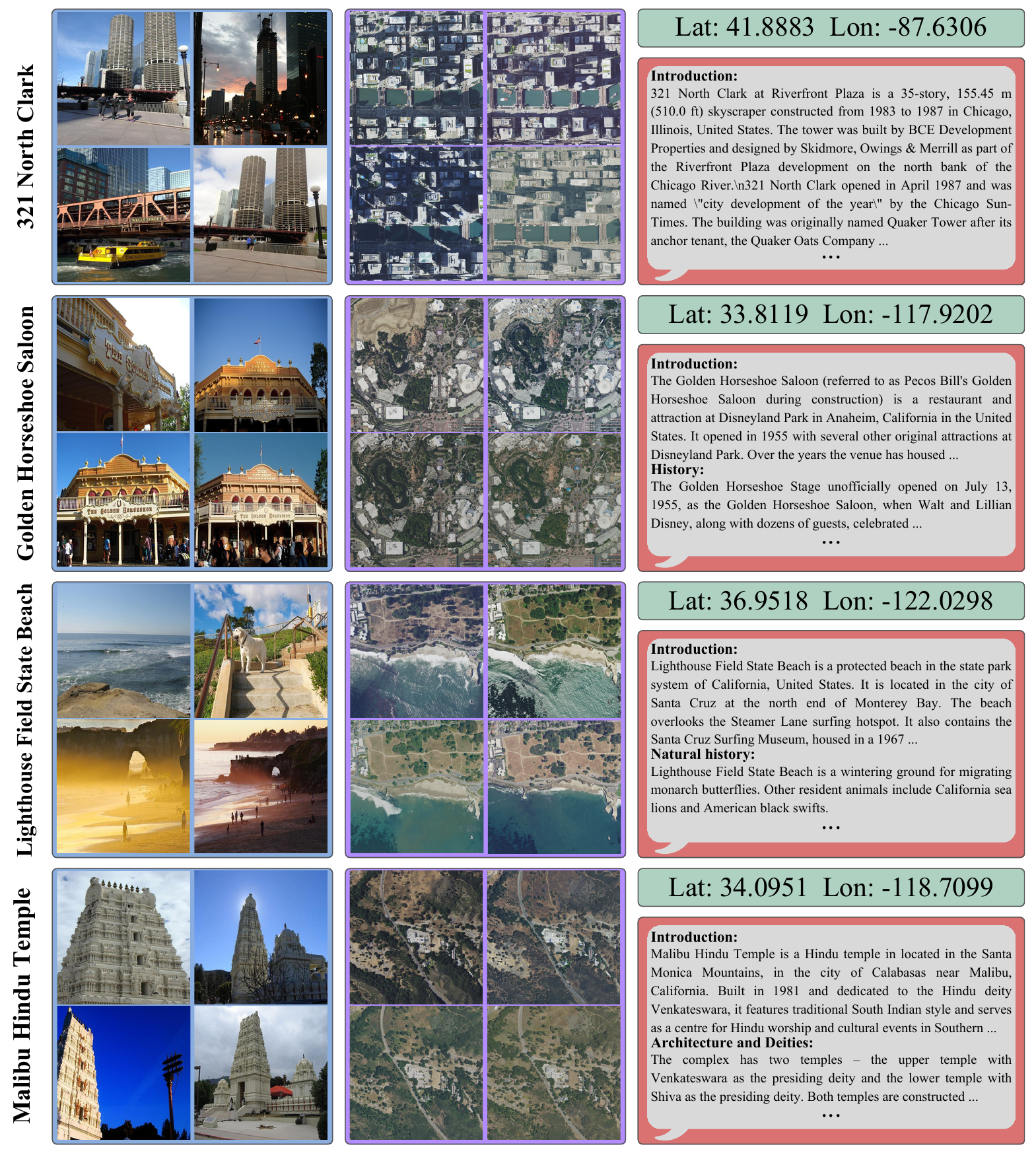}
    \caption{Additional examples from the \dname dataset. We illustrate the diversity in the dataset by randomly sampling landmarks. We show sample ground and satellite views, as well as the exact GPS location and parts of the textual descriptions.}
    \label{fig:more_examples_2}
\end{figure*}

\begin{figure*}
    \centering
    \includegraphics[width=\textwidth]{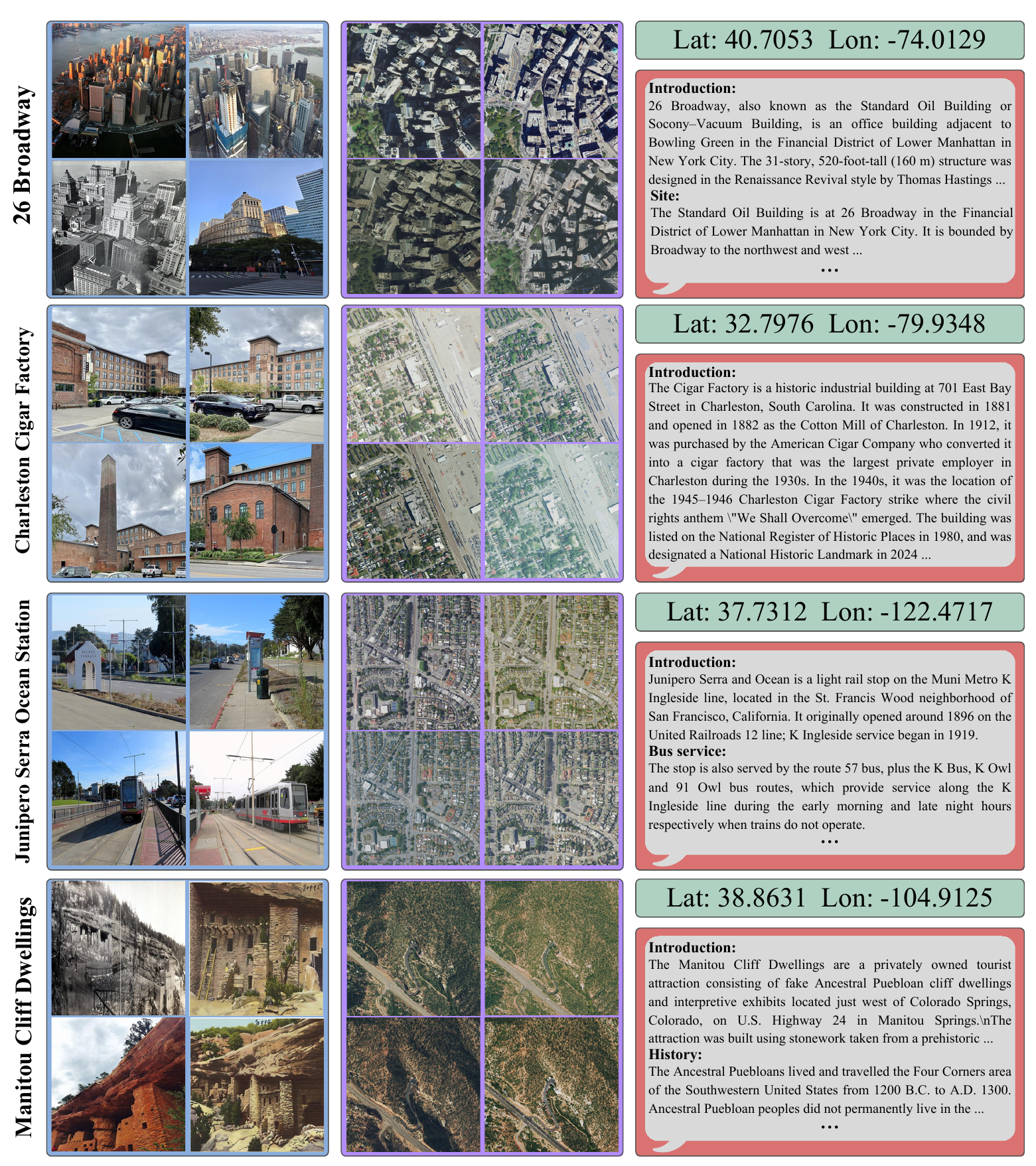}
    \caption{Additional examples from the \dname dataset. We illustrate the diversity in the dataset by randomly sampling landmarks. We show sample ground and satellite views, as well as the exact GPS location and parts of the textual descriptions.}
    \label{fig:more_examples_3}
\end{figure*}

\begin{figure*}
    \centering
    \includegraphics[width=\textwidth]{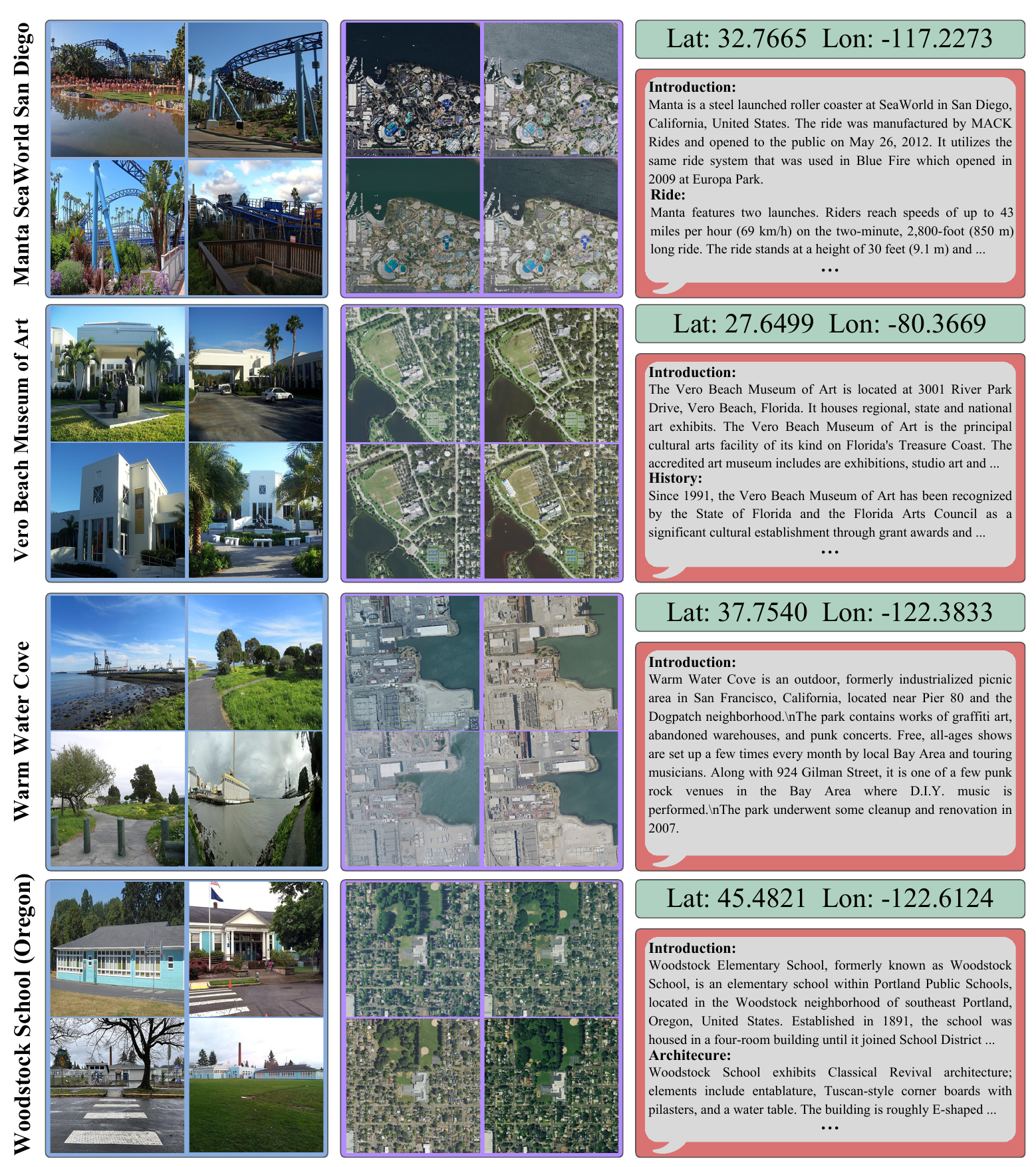}
    \caption{Additional examples from the \dname dataset. We illustrate the diversity in the dataset by randomly sampling landmarks. We show sample ground and satellite views, as well as the exact GPS location and parts of the textual descriptions.}
    \label{fig:more_examples_4}
\end{figure*}

For the ``Red Rocks Amphitheatre", an asphalt parking lot was added between the times when images 1 and 2 were captured. Additionally, the top of the Amphitheatre is now covered by a greyish roof, which is also visible in ground-level images found through a web search for the landmark.
The \dname is extremely diverse in terms of geography, textual, visual, and temporal complexity. We show additional examples of our dataset in Figures \ref{fig:more_examples_1}-\ref{fig:more_examples_4}. \dname offers the possibility to train and evaluate models in a unified framework that more realistically reflects our world.

\section{VLM Data Processing}
\label{appendix:vlm_filtering}

In Fig.~\ref{fig:vlm_processing_error}, we show the splitting process used to create the subset with only outdoor images (83\% of the ground images classified as outdoors, as mentioned in Sec.\ 3. We use a Vision Language Model (\texttt{llava-hf/llava-1.5-7b-hf})~\cite{llava} 
and prompt the following: ``USER: $<$image$>$ Does this image contain the outdoors view of a building, place, or landmark?\textbackslash n ASSISTANT:". The VLM's answering window is made very small, such that the VLM only replies by ``Yes" (outdoors) or ``No" (indoors). In doing so, we remove all images where there are no obvious cues in the image that depict the landmark seen from the exterior. After all images are processed, we sample $1000$ ground views randomly and manually inspect them for soundness. We find that $8.2\%$ are wrongly classified, with $3.7\%$ wrongly categorised as indoors, and $4.5\%$ wrongly categorised as outdoors.

However, the $3.7\%$ that are wrongly classified as indoors and should have been marked as outdoors are actually relevant to filter out, since they are often close-up, natural images that do not provide any particular information about the landmark, and would therefore not contribute during training had they been correctly labelled. This is reflected in the bottom row of Fig.~\ref{fig:vlm_processing_error}, where the VLM classifies a close view of a building or natural landscapes where it cannot see any noticeable landmark, as ``No" landmark.

For the images from an indoor setting that are incorrectly classified as ``outdoors", many wrong conclusions from the model stem from basketball courts, malls, or the interior of large buildings that share similar architecture as outdoor landmarks (respectively seen in the top row of Fig.~\ref{fig:vlm_processing_error}: images one, two, and five). Some scanned documents and paintings of buildings are also falsely classified as outdoor images. Upon inspection, we conclude that the number of wrong classifications is acceptably low. The results for models trained on the outdoors subset and presented in Tab.~\ref{tab:all_model_experiments} 
also demonstrate the relevance of employing such a filtration process as a pre-processing clean-up of the raw \dname dataset.


\end{document}